%% file: text_fidelity.tex
\documentclass{article}

\usepackage[preprint]{neurips_data_2024}

\usepackage[utf8]{inputenc} %
\usepackage[T1]{fontenc}    %
\usepackage{hyperref}       %
\usepackage{url}            %
\usepackage{booktabs}       %
\usepackage{amsfonts}       %
\usepackage{nicefrac}       %
\usepackage{microtype}      %
\usepackage{amsmath, amssymb} 
\usepackage{graphicx}
\usepackage[table]{xcolor}
\usepackage{wrapfig,lipsum,booktabs}
\usepackage{tcolorbox}
\usepackage{color, colortbl}
\usepackage{duckuments} %
\usepackage[export]{adjustbox}

\definecolor{mygreen}{RGB}{34, 139, 34}
\definecolor{myblue}{RGB}{36, 130, 200}
\definecolor{myyellow}{RGB}{180, 155, 100}

\newcolumntype{H}{@{}>{\lrbox0}l<{\endlrbox}@{}}

\usepackage{xspace}
\newcommand{\ours}{\textsc{TypeScore}\xspace}
\newcommand{\oursdata}{\textsc{TypeInst}\xspace}

\NewDocumentCommand{\yz}
{ mO{} }{\textcolor{blue}{\textsuperscript{\textit{Yizhe}}\textsf{\textbf{\small[#1]}}}}

\NewDocumentCommand{\rx}
{ mO{} }{\textcolor{cyan}{\textsuperscript{\textit{Ruixiang}}\textsf{\textbf{\small[#1]}}}}

\def\clipscore{CLIPScore~}
\def\dalle{DALL-E 3~}
\def\mj{MidJourney~}
\def\ideogram{ideogram~}
\def\sd{Stable Diffusion~}

\title{\ours: A Text Fidelity Metric for Text-to-Image Generative Models}

\author{%
  Georgia Gabriela Sampaio, Ruixiang Zhang, Shuangfei Zhai, Jiatao Gu, \\\textbf{Josh Susskind, Navdeep Jaitly, Yizhe Zhang}\\
  Apple\\
  \texttt{ \{gsamp, ruixiangz, szhai, jgu32, jsusskind, ndjaitly, yizzhang\}@apple.com} \\
}

\begin{document}

\maketitle
\begin{abstract}
Evaluating text-to-image generative models remains a challenge, despite the remarkable progress being made in their overall performances.
While existing metrics like \clipscore work for coarse evaluations, they lack the sensitivity to distinguish finer differences as model performance rapidly improves. In this work, we focus on the text rendering aspect of these models, which provides a lens for evaluating a generative model's fine-grained instruction-following capabilities. To this end, we introduce a new evaluation framework called \ours to sensitively assess a model's ability to generate images with high-fidelity embedded text by following precise instructions. We argue that this text generation capability serves as a proxy for general instruction-following ability in image synthesis. \ours uses an additional image description model and leverages an ensemble dissimilarity measure between the original and extracted text to evaluate the fidelity of the rendered text.
Our proposed metric demonstrates greater resolution than \clipscore to differentiate popular image generation models across a range of instructions with diverse text styles.
Our study also evaluates how well these vision-language models~(VLMs) adhere to stylistic instructions, disentangling style evaluation from embedded-text fidelity. 
Through human evaluation studies, we quantitatively meta-evaluate the effectiveness of the metric. Comprehensive analysis is conducted to explore factors such as text length, captioning models, and current progress towards human parity on this task. The framework provides insights into remaining gaps in instruction-following for image generation with embedded text. \footnote{Code and data will be made publicly available soon to facilitate future work on this challenging problem.}
\end{abstract}

\section{Introduction}

Image generation models have seen significant advancements in recent years, producing high-quality and diverse synthetic images. Notable examples include \dalle \citep{ramesh2021zeroshot}, \ideogram \citep{ideogram}, \sd \citep{rombach2022high}, \mj \citep{midjourney2022}, Imagen \citep{imagen}, Dream \citep{dream} and Adobe Firefly \citep{adobefirefly}. However, even these high-quality image generation models often struggle to generate images with specific text embedded within them. This lack of \textit{embedded-text fidelity} can take the form of typos, repeated or missing characters and words, extraneous characters, and unreadable glyphs.

Unfortunately, current metrics such as \clipscore \citep{Hessel2021CLIPScoreAR} are unsuitable for measuring \textit{embedded-text fidelity}. These metrics work well when there is a large performance gap between models~\citep{DBLP:journals/corr/abs-2404-03539}, but are not sensitive to more nuanced improvements as the quality of generation improves \citep{chen2023understanding} because they rely on image embeddings that can lose the fine-grained details required to detect nuanced differences. 
As the quality of image generation has improved, there is a growing need for new metrics specifically designed to evaluate these models' instruction-following capability in a microscopic manner.

\begin{figure}[h]
\centering
\includegraphics[width=\textwidth]{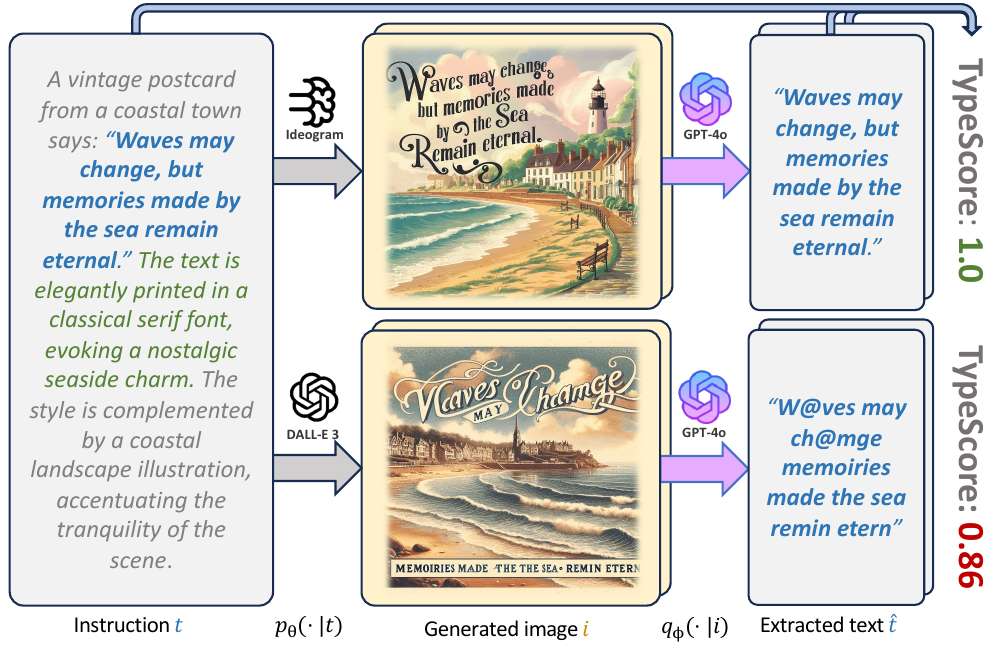}
\caption{When assessing target image generation models $p_\theta$, we provide the model with a set of instructions. These instructions prompt the model to create a set of \textcolor{myyellow}{images} $i$ based on \textcolor{myblue}{specified quoted text} within a \textcolor{mygreen}{particular style}, alongside some \textcolor{gray}{contextual information}. We then use a vision-language model $q_\phi$ (\textit{e.g.} GPT-4o) to extract the text from the \textcolor{myyellow}{generated images}, and compute the similarity score between the \textcolor{myblue}{generated text} $\hat{t}$ and the \textcolor{myblue}{original quote} $t$. \ours is calculated by averaging the scores obtained from multiple image generations. Common text-image alignment metrics such as \clipscore produce indistinguishable results for both image generation models under this prompt.}
\label{fig:typescore}
\end{figure}

We aim to bridge this gap by introducing a new evaluation metric that probes the performance differences among competitive image generation models. We propose \ours, an evaluation metric designed to assess the fidelity of embedded text in generated images~(Figure~\ref{fig:typescore}). \ours offers a precise and nuanced assessment of embedded-text fidelity, incorporating key factors such as legibility and accuracy. Style can be a confounding factor in assessing embedded-text fidelity. We present effective methods to ground the generation with rich contextual and style instructions to create a controlled environment that minimizes the confounding aspects like text font, typeface, and aesthetic integration.

The evaluation framework for \ours includes a probing instruction dataset (\oursdata) of 118 text-embedded image generation instructions with diverse requirements of styles, text formatting, and length. 
To meta-evaluate different variants of \ours and compare \ours with CLIPScore, we crowd-sourced annotations of human preferences on \textit{text fidelity}, \textit{style fidelity} and \textit{overall preference}, over 590 pairs of generated images, and score each metric according to their alignment with the human preferences. 
The resulting \ours is an ensemble score over multiple dissimilarity metrics, and it aligns significantly better with human preference than CLIPScore. 

We show that \ours has the sensitivity required to differentiate between the embedded-text fidelity of several state-of-the-art image generation models while CLIPScore is unable to detect these subtle differences. Interestingly, we found that models with higher \ours were also ranked higher in our annotated preferences for style-following and general instruction-following, indicating this metric can extrapolate to serve as a proxy for the model's general ``instruction-following'' ability. 
We further discuss the impact of text extraction models, instruction length, and re-captioning on the sensitivity of \ours. 

While our work focuses on text fidelity evaluation in text-to-image generative models, we argue that generating faithful embedded text is actually a cutting-edge challenge that probes the most sophisticated capabilities of image generation models, thus it can potentially serve as a broader indicator of a model’s fine-grained control and instruction-following abilities, as evidenced by the correlation between text fidelity scores and overall model performance in our evaluations.

Specifically, our key contributions are:
\begin{enumerate}
    \item We introduce a new metric for evaluating image generation quality for rendering embedded text, highlighting the blind spots of CLIPScore. Our metric demonstrates higher sensitivity compared to existing metrics like \clipscore for high-quality image generation models.
    \item We provide an evaluation framework, including an instruction dataset \oursdata with detailed and diverse image style descriptions and quoted text.
    \item We perform large-scale human evaluation to quantitatively meta-evaluate the proposed metrics.
    \item We provide insights into the current capabilities of image generation models and their progress towards achieving human-like proficiency in generating images with embedded text.
    \item We conduct thorough ablation studies and analyses to understand the effects of text length, the use of different visual-language-models (VLMs) for captioning, and the impact of re-captioning.
\end{enumerate}

\section{Related Work}

\paragraph{Image Generation with Embeded Text}
Advances in image generation models have significantly improved the quality of synthetic images. Models such as DALL-E 3 \citep{ramesh2021zeroshot}, Stable Diffusion 3 \citep{rombach2022high}, ideogram \citep{liu2023ideogram}, and MidJourney \citep{midjourney2022} have shown remarkable progress in producing diverse and high-quality visuals. Despite their success and rapid improvement, these models still encounter challenges in generating text with high fidelity, often producing text with typos, repeated or missing characters, and extraneous glyphs. Many methods have been proposed to improve the fidelity of embedded text in the generation. TextDiffuser \citep{chen2023textdiffuser} addresses these issues by using a two-stage process: first, a Transformer model generates the layout of keywords from text prompts, and then diffusion models generate images conditioned on these layouts. 
TextDiffuser-2 \citep{chen2023textdiffuser2} further enhances text rendering by integrating large language models for layout planning and text encoding, enabling more flexible and diverse text generation.
AnyText \citep{tuo2023anytext} takes a different approach by focusing on multilingual visual text generation and editing, leveraging a diffusion pipeline to first mask the image and then employ an Optical Character Recognition (OCR) model to encode stroke data as embeddings to generate texts that can integrate with the background. 
However, these models typically involve multiple components, and generating text with both high fidelity and aesthetic and natural style remains challenging, as high-fidelity text generation frequently sacrifices rendering quality and artistic value.

\paragraph{Text Fidelity Evaluation Metrics and Datasets}
Traditional text fidelity metrics such as CIDEr \citep{vedantam2015cider}, SPICE \citep{anderson2016spice}, and BLEU \citep{papineni2002bleu} have been widely used for evaluating image captions. CIDEr focuses on consensus in large datasets, SPICE uses scene graph structures for more detailed semantic evaluation, and BLEU measures n-gram precision against reference texts. While these metrics have been foundational, they sometimes fall short in capturing the holistic meaning. Our approach is based on these standard word-overlapping-based methods. \citet{wang2021faier} introduced the FAIEr metric, designed to assess both the fidelity and adequacy of image captions. FAIEr employs a visual scene graph to bridge the image and text modalities, leveraging it as a criterion for fidelity evaluation and guiding adequacy assessment through reference captions. 

When testing the instruction-following ability of image generation models, existing text-image alignment metrics like CLIPScore \citep{Hessel2021CLIPScoreAR} evaluate images by computing the cosine similarity between image and text embeddings of the instruction. \citet{xu2023imagereward} introduced ``ImageReward,'' a model designed to learn and evaluate human preferences for text-to-image generation, aiming to improve the alignment of generated images with human aesthetic and contextual preferences through systematic annotation and reward feedback mechanisms. 
\citet{lee2024holistic} proposes a comprehensive evaluation framework for assessing text-to-image models, addressing the limitations of existing metrics that focus narrowly on specific aspects. The framework integrates multiple dimensions, including fidelity, diversity, relevance, and creativity, to provide a more balanced assessment. The authors introduce a new benchmark combining automatic metrics and human evaluations to improve understanding and performance of these models.
\citet{somepalli2024measuring} focuses on evaluating and understanding the stylistic attributes of images generated by diffusion models, and the proposed model demonstrates superior performance in style retrieval tasks compared to previous methods. Our approach differs from them as we focus on testing the fidelity of the embeded text in the generated image. 
TextDiffuser \citep{chen2023textdiffuser} also introduces the MARIO-10M dataset and MARIO-Eval benchmark to enhance and evaluate text rendering quality. They evaluate the generation by using OCR to extract the text. Our work extensively evaluates different options for text description models and shows that OCR can yield suboptimal extraction results when the generated image is stylish.

\section{Text Fidelity Assessment}

Given the image generation instructions \( t \sim \mathcal{T} \), let's assume an image generation model $p_\theta$ produces a corresponding image \( i \sim p_\theta(\cdot|t)\).
A model exhibiting good \textit{instruction-following} ability would create an image \( i \) that: 1) conforms with all the information provided in the instruction \( t \), and 2) refrains from generating extraneous elements beyond the given instruction.
Therefore, assessing the instruction-following capability of an image generation model can be perceived as evaluating the mutual information \( \mathbf{MI}(i, t) \) of the joint distribution, which is defined as
\begin{align}
\mathbf{MI}_\theta(i, t) = \mathbb{E}_{t \sim \mathcal{T}, i \sim p_\theta(\cdot|t)} \log \frac{p(i, t)}{p(i) p(t)} 
\label{eq:mi}
\end{align}

However, directly evaluating the instruction-following capability would be challenging as there are numerous ways to describe an image. To probe the general instruction-following capability of an image generation model, we study a more \textit{controlled} problem of the \textit{embedded-text fidelity assessment} task, which evaluates how faithful an image generation model follows the instruction to render a specific piece of text in a certain style. As the evaluation focuses on the embedded-text fidelity, this yields a clear evaluation metric.  

Consider a dataset \(\mathcal{D}\) containing \(N\) image generation instructions, where each instruction includes a quoted text \(t\). Each image generation model is tasked with generating an set of images \(\{i_1,\cdots,i_N\}\), based on the instructions.
We investigate the following: how accurately are the rendered text in the images compared to the quoted text \(t\) from the instruction?

\subsection{\ours: A Text Fidelity Evaluation Framework for Image Generation Models}
Suppose we have a reverse model $q_\phi$ that can predict the instruction $t$ from the image $i$, from \eqref{eq:mi} (see Appendix \ref{app:proof} for proof),
\begin{align}
    \mathbf{MI}_\theta(i, t) \geq \mathbb{E}_{t \sim \mathcal{T}} \mathbb{E}_{i \sim p_\theta(\cdot|t)} \log q_\phi(t|i) \triangleq \mathcal{L}_{\mathbf{MI}}(\theta;\phi).
\end{align}

This suggests that instead of directly estimating $\mathbf{MI}$, we can potentially use an image description model $q_\phi$ to calculate a lower bound proxy $\mathcal{L}_{\mathbf{MI}}(\theta;\phi)$ of $\mathbf{MI}$. Due to the rapid advancement of Vision-Language Models (VLMs), obtaining this $q_\phi$ becomes more convenient and $q_\phi$ can be good off-the-shelf zero-shot posterior approximators in this context.
To evaluate the image generation model's capability in generating accurate text based on instructions, we introduce an evaluation framework that leverages an image description model $q_\phi$.
Practically, we can ask $q_\phi$ to either calculate the likelihood of $t$, or to generate an estimate $\hat{t}$, using a similarity measure $S(t, \hat{t})$ as the metric when $q_\phi$ does not produce a likelihood.

In the following, we discuss our evaluation framework, which consists of a dataset of diverse instructions, an image description model to extract text from the images, and an ensemble score to measure the difference between the extracted text and the original quoted text.

\subsection{\oursdata Dataset}

\begin{figure}[h]
\centering
\includegraphics[width=\textwidth]{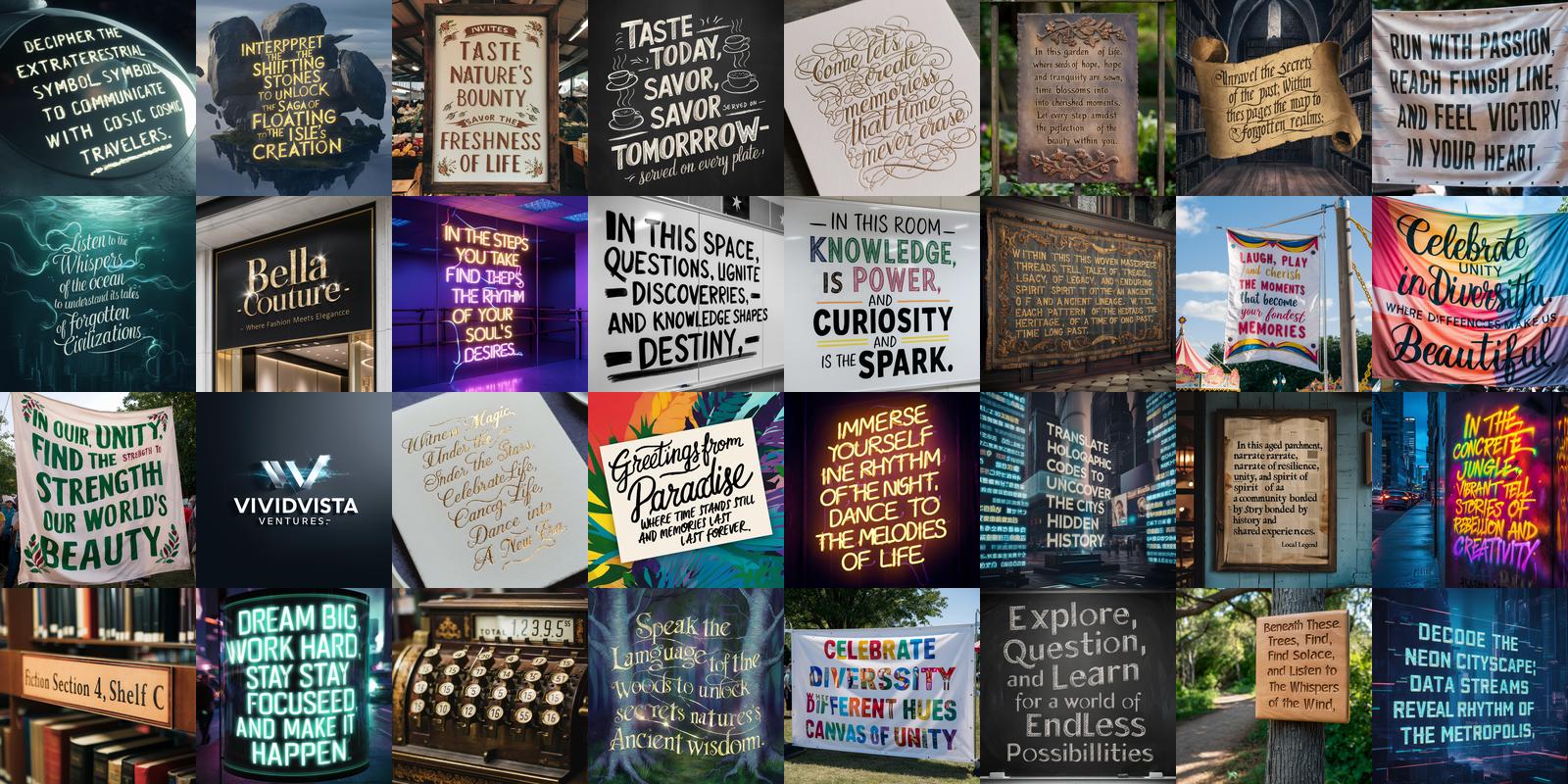}
\caption{Sampled generations of ideogram using the instructions from \oursdata.}
\label{fig:collage}
\end{figure}

We create the dataset of text instructions, \oursdata, using GPT-3.5 \citep{brown2020language} by prompting the model with basic text and style elements written by the authors. Following the Magic Prompt approach in ideogram \citep{ideogram}, the GPT-3.5 model was prompted to enhance and recaption the initial raw descriptions in three iterations, resulting in rich instructions that offer comprehensive details about both the image and the text style. The text to be rendered is in between quotes. See Figure~\ref{fig:collage} for sampled generations using the instructions from \oursdata.

\oursdata dataset comprises 118 instructions across various scenarios for evaluation. The average number of words per instruction is 33.94, while the average number of words in the quote is 11.78.
\oursdata covers a broad range of domains and subjects (see Appendix \ref{dataset_composition} for the composition of \oursdata) such as celebratory milestones, futuristic adventures, urban life, cozy settings, inspirational messages, historical themes, cultural celebrations, natural landscapes, educational environments, and artistic expressions. It also includes practical text instances beyond English characters, like addresses, digits, acronyms, and logos. This variety encompasses different styles, fonts, and contexts, providing a comprehensive resource for evaluation.

\subsection{Image Description Methods}
With the generated images from each tested model, we employ an image description model $q_\phi$ to extract the rendered text from these images. We compared the performance of two different classes of image captioning models to extract captions: OCR \citep{ocr}, and Vision-language Models (VLMs).
The VLMs were instructed to extract only the rendered text while preserving any typos or errors. We also ask the model to use ``@'' tokens to represent any glyphs or extraneous symbols that cannot be reasonably interpreted to match any English character (Figure~\ref{fig:typescore}). The full prompt is provided in below:

\begin{center}
\begin{tcolorbox}[width=0.99\textwidth]
\textit{Identify the main text contained in this image, and output it between quotes, 
                without correcting any typos or issues you may encounter. \textbf{Do not output anything else.}
}
\end{tcolorbox}
\end{center}

To compare these models, we ask human annotators to extract the rendered text from 590 generated images (see Appendix \ref{app:human_eval} for details and the screenshot of the annotation interface). Using this human extraction as ground truth $\hat{t}_{\text{oracle}}$, we can compute the extraction accuracy using Normalized Edit distance (NED) \citep{normalized_edit_distance} for each of the $q_\phi$.
Note that as some of the $q_\phi$ tend to auto-correct the extracted text, the similarity score between original text $t$ and the extracted text $\hat{t} \sim q_\phi(\cdot|i)$ can be even higher than the similarity between $t$ and $\hat{t}_{\text{oracle}}$. Therefore, the automatic score of similarity cannot be used to judge on which $q_\phi$ is more accurate. The comparison of the $q_\phi$ is provided in Table \ref{tab:vlms}.

\begin{wraptable}{r}{0.48\textwidth}
\centering
\rowcolors{2}{gray!25}{white} %
\begin{tabular}{lc}
    \toprule
    \rowcolor{white} \textbf{Models} & NED($\hat{t}_{\text{oracle}}, \hat{t}_{q_\phi}$) ($\downarrow$) \\
    \midrule
    {OCR} & 0.650 $\pm${0.032} \\
    LLaVa-NeXT & 0.618 $\pm${0.023} \\    
    GPT-4v & 0.340 $\pm${0.029} \\    
    OCR + GPT-4o & 0.355 $\pm${0.033} \\
    GPT-4o & \textbf{0.315 $\pm${0.030}} \\
    \bottomrule
\end{tabular}
\caption{The Normalized Edit Distance between human oracle extraction and each $q_\phi$'s extraction. GPT-4o yields the highest alignment with human oracle extraction.}
\label{tab:vlms}
\end{wraptable}

\begin{figure}[ht!]
\centering
\includegraphics[width=0.85\textwidth]{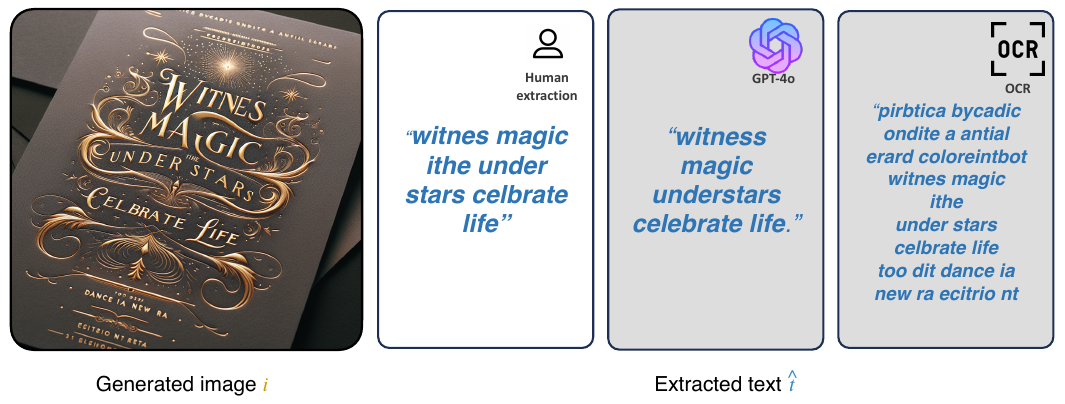}
\caption{When extracting text, OCR tends to introduce errors, while VLMs tend to autocorrect existing errors in the rendered text.  }
\label{fig:ocr}
\end{figure}

We observed that OCR underestimates the \ours by failing to identify the main text and introducing extraneous characters and symbols from glyphs. Conversely, VLMs tend to overestimate \ours by fixing typos and incorrect word ordering \ref{fig:ocr}. However, with a careful prompt tuning, this overestimation issue can be alleviated. In practice, we used GPT-4o \citep{gpt4o} as in our experiments it leads to the best extraction accuracy comparing to other alternatives, including OCR, GPT-4v and LLaVa-NeXT \citep{liu2024llavanext}. We combined OCR and GPT-4o (\textbf{OCR+GPT-4o}) by feeding the OCR output into GPT-4o and prompting the model to discern which portions of the OCR output were extraneous to the main text, thereby preserving the original typos from OCR while utilizing GPT-4o's strength in accurately identifying the main text. Combining OCR and GPT-4o fails to outperform GPT-4o. The prompt used to refine the OCR output with GPT-4o is provided in Appendix \ref{gpt4o_ocr_prompt}.

\subsection{Scoring Mechanism}
Image generation models may struggle to accurately render text due to various factors, such as typos (missing, repeated, or unnecessary characters), errors in word order or repetition, and even generating unintelligible text or no text at all. 
Given the extracted text $\hat{t}$, we explored a set of dissimilarity metrics that cover different error factors.

\begin{itemize}
    \item \textbf{Normalized Edit Distance} \citep{normalized_edit_distance} measures the ratio of edit distance to the average of the length of both strings, providing a normalized measure of dissimilarity between two strings. These metrics were chosen for their effectiveness in quantifying deviations at the character level, enabling the identification of typographical anomalies such as misspellings, character omissions, and insertions.  
    \item \textbf{BLEU} \citep{bleu} evaluates the precision of machine-generated translations by comparing them to a reference translation based on exact word matches. BLEU-1 was selected to evaluate fidelity at the word level, enabling the detection of discrepancies in word choice, repetition, and omission. 
    \item \textbf{Character-BLEU} evaluates the precision of machine-generated translations by comparing them to a reference translation based on exact \textit{character matches}, rather than entire words. 
    \item \textbf{Normalized Longest Common Subsequence} (NLCS) \citep{longest_common_subsequence} measures the length of the longest common subsequence between two strings, normalized by dividing by the length of the longer string, providing a similarity measure between the two strings. These metrics were selected to assess text completeness by quantifying the extent to which the generated text aligns with the original text, thereby providing insights into both word order fidelity and text integrity.
    \item \textbf{Smith Waterman} \citep{smith_waterman} is a local sequence alignment algorithm used to identify the optimal local alignment between two sequences by scoring matches, mismatches, and gaps. 
    \item \textbf{Ensemble Score} comprises a subset of the aforementioned distance metrics.
\end{itemize}
Alternatively, we could compute the likelihood $q_\phi(t|i)$ for the original quote $t$.
Directly calculating such likelihood by extracting all logits from the GPT-4 API is challenging because the API does not natively support log probability evaluation of input tokens. It only supports log probability evaluations for up to the top $20$ generated tokens, making it difficult to compute likelihoods in a reasonable and reliable manner.

\subsection{Evaluated Baseline Models}
\label{sec:baseline}
We evaluated four image generation models that showcase the SOTA text rendering capabilities of current VLMs: DALL-E 3 \citep{ramesh2021zeroshot}, Stable Diffusion XL \citep{podell2023sdxl}, Stable Diffusion 3 \citep{sd3}, and ideogram \citep{ideogram}. These models were used only for inference in their default configurations to generate images according to the given instructions. We use 8x Nvidia A100 GPUs for all the experiments.

\section{Meta-evaluation of the \ours Variants}
\paragraph{Human annotation} To meta-evaluate different variants of \ours and compare our method with CLIPScore, 
we further performed pairwise human evaluation task on 472 pairs of images generations from DALL-E 3, ideogram, \sd and \sd3 on our internal crowd-source annotation platform. 
Each pair of generated images was evaluated by 3 to 5 judges, presented in random order with in-context examples and detailed instructions of the requirements and evaluation aspects. We aimed for at least 60\% agreement: if 2 out of 3 judges concurred on a top answer, the evaluation concluded. Otherwise, up to 2 additional judges were consulted to achieve the desired agreement. The judges were instructed to assess on three tasks using a 3-point Likert-like scale:
\begin{enumerate}
    \item \textbf{Text fidelity:} \textit{In which image does content of the rendered text better align with the original quote?}
    \item \textbf{Style fidelity:} \textit{In which image does the style better align with the style description in the instruction?}
    \item \textbf{Overall preference:} \textit{Considering the content of the rendered text, alignment with the instruction and aesthetic value, which image better aligns with the given instruction?}
\end{enumerate}

Further details, including the human evaluation template used, hourly rate of the evaluation task, and inter-rater agreement analysis, are provided in Appendix ~\ref{app:human_eval}. 

\paragraph{Results}
Based on the \textbf{Text fidelity} annotation, we compare different variants of \ours via the meta-metrics of \textit{alignment accuracy}, which is computed as
\begin{align}
    \text{Alignment Accuracy} = \frac{|(\ours(\theta)>\ours(\theta')) \cap \text{Human prefer} \,\, \theta \,\,  \text{over} \,\, \theta' | }{|\text{All annotated pairs}|}, \nonumber
\end{align}
where $|\cdot|$ denotes the cardinality. 
The alignment accuracy indicates the percentage of pairs where automatic metrics and human preferences agree. Tied pairs are excluded from the calculation. This metric is linked to Percent Rank Violation (PRV) \citep{li24simpler}, which evaluates the ranking violation of the metrics against the oracle preference. 
In this way, we calculate the ranking violation of the metrics for each pair of models using human-annotated data, then compute the aggregated means. The results are presented in Table~\ref{tab:accuracies}.
We found that \ours (character-BLEU), \ours (Smith Waterman) and \ours (NED) generally exhibit better alignment with human judgement of text fidelity, style fidelity and overall preference compared to other distance metrics. After normalizing each distance metric to $[0,1]$, combining NED, Smith-Waterman and NLCS through mean pooling effectively leverages their strengths, resulting in robust, generalizable and high alignment accuracy. We refer to the resulting ensemble methods as \ours in the subsequent discussion. \ours consistently outperforms CLIPScore in this meta-evaluation. Further examples are provided in the Appendix \ref{text_fidelity}, illustrating how \ours can discern subtle differences in rendered text. We also noted that in cases where there is a significant quality gap between pairs, both CLIPScore and \ours can accurately detect the difference. Yet, when the quality gap is narrower (\textit{e.g.}, DALL-E vs Ideogram), CLIPScore consistently struggles to rank them correctly, whereas \ours remains sensitive. 

\begin{table*}[ht!]
\centering
\rowcolors{2}{gray!25}{white} %
\begin{tabular}{lccc}
    \toprule

    \rowcolor{white} \textbf{Alignment Accuracy} & \textbf{Text Fidelity}($\uparrow$) & \textbf{Style Fidelity}($\uparrow$) & \textbf{Overall Prefer.}($\uparrow$)\\
    \midrule
    {CLIPScore} \citep{Hessel2021CLIPScoreAR} & $66.3\% \pm{0.6\%}$ & $58.3\% \pm{0.9\%}$ & $65.8\% \pm 0.7\%$ \\
    \midrule
    {\ours (NED)} & $69.0\% \pm{0.5\%}$ & $61.7\% \pm{0.5\%}$ & $68.6\% \pm{0.3\%}$ \\
    {\ours (BLEU)} & $69.3\% \pm {0.4\%}$ & $60.6\% \pm{0.9\%}$ & $69.2\% \pm{0.5\%}$ \\
    {\ours (character-BLEU)} & $\textbf{71.5}\% \pm{0.8\%}$ & $62.0\% \pm {0.5\%}$ & $71.1\% \pm{0.7\%}$ \\
    {\ours (NLCS)} & $67.8\% \pm{0.6\%}$ & $59.3\% \pm {0.9\%}$ & $67.7\% \pm{0.8\%}$ \\
    {\ours (Smith Waterman)} & $69.3\% \pm{0.5\%}$ & $\textbf{63.9}\% \pm {0.7\%}$ & $69.2\% \pm{0.4\%}$\\
    \midrule
    {\ours (Ensemble Score)} & $71.1\% \pm {0.5\%}$ & $62.2\% \pm {0.7\%}$ & $\textbf{71.3}\% \pm{0.3\%}$ \\
    \bottomrule
\end{tabular}
\caption{Alignment Accuracy of CLIPScore and \ours variants based on human preference, w.r.t text fidelity, style fidelity and overall preference. GPT-4o is used for the text extraction. \ours aligns better with human preference of text fidelity, style fidelity and overall preference. Averaging the three columns, \ours (Ensemble Score) yields the robust and highest alignment accuracy.}
\label{tab:accuracies}
\end{table*}

\paragraph{Extrapolation property of \ours} Interestingly, as shown in Table~\ref{tab:accuracies}, our approach also demonstrates good alignment with \textbf{Style fidelity} and \textbf{Overall preference}, while CLIPScore falls short. This suggests that text fidelity can be closely associated with the evaluation of general instruction-following ability of an image generation model. Therefore, we can use \ours to \textit{probe} the image generation model's capability to follow instructions, particularly if they are not specifically tailored to optimize rendered text fidelity. 

\paragraph{\ours sensitivity to Text Length} 

We found that \ours is robust to variations in input text length, yielding consistent performance regardless of the length of the instruction or quoted text. We validated this robustness by computing the Pearson correlation \citep{pearson_correlation_1, pearson_correlation_2} between \ours and varying input text lengths, and found no significant correlation across several image generation models. This indicates that the \ours is stable and reliable across texts of varying lengths. See Appendix \ref{text_length} for more details. 

\paragraph{\ours sensitivity to Recaptioning} 

We found that recaptioning the input image description by adding more stylistic details and contextual information helps control the text fidelity evaluation of the rendered text.
 This improvement is reflected in higher \ours and slightly lower variance, demonstrating that incorporating richer stylistic nuances contributes to a more controlled setting for text fidelity evaluation. We measured this by prompting ideogram to generate images using an augmented input caption and comparing the scores of the resulting images. See Appendix \ref{recaptioning} for more details.

\paragraph{TYPESCORE sensitivity to Text Extraction Method} 
We conducted a comprehensive analysis to evaluate TYPESCORE's robustness across different text extraction methods. While our primary quantitative analyses utilize GPT-4o due to its highest alignment with human extractions, we found that TYPESCORE maintains consistent performance across different text extraction approaches. To demonstrate this, we evaluated the alignment accuracy using LLaVA-NEXT, a more efficient alternative to GPT-4o. Despite LLaVA-NEXT showing lower alignment with human oracle extractions ($0.618$ mean normalized edit distance) compared to GPT-4o, TYPESCORE consistently outperforms CLIPScore in aligning with human preferences across all evaluation aspects (Table~\ref{tab:llava_accuracies}).

\begin{table*}[ht!]
\centering
\rowcolors{2}{gray!25}{white}
\begin{tabular}{lccc}
    \toprule
    \rowcolor{white} \textbf{Alignment Accuracy (LLaVA):} & \textbf{Text Fidelity}($\uparrow$) & \textbf{Style Fidelity}($\uparrow$) & \textbf{Overall Prefer.}($\uparrow$)\\
    \midrule
    {CLIPScore} & $38.4\% \pm{0.8\%}$ & $36.1\% \pm{0.5\%}$ & $38.0\% \pm 0.5\%$ \\
    \midrule
    {TYPESCORE (NED)} & $39.9\% \pm{0.4\%}$ & $37.1\% \pm{0.9\%}$ & $39.9\% \pm{0.7\%}$ \\
    {TYPESCORE (BLEU)} & $38.9\% \pm{0.5\%}$ & $\textbf{37.7}\% \pm{0.6\%}$ & $38.9\% \pm{0.5\%}$ \\
    {TYPESCORE (character-BLEU)} & $\textbf{40.2}\% \pm{0.6\%}$ & $37.5\% \pm{0.9\%}$ & $\textbf{40.2}\% \pm{0.5\%}$ \\
    {TYPESCORE (NLCS)} & $38.9\% \pm{0.3\%}$ & $36.5\% \pm{0.8\%}$ & $38.7\% \pm{0.6\%}$ \\
    {TYPESCORE (Smith Waterman)} & $39.9\% \pm{0.4\%}$ & $\textbf{37.7}\% \pm{0.7\%}$ & $40.0\% \pm{0.7\%}$ \\
    \midrule
    {TYPESCORE (Ensemble Score)} & $39.0\% \pm{0.4\%}$ & $37.0\% \pm{0.7\%}$ & $39.3\% \pm{0.6\%}$ \\
    \bottomrule
\end{tabular}
\caption{Alignment accuracy using LLaVA-NEXT as the text extraction model. Despite using a less accurate text extraction model, TYPESCORE maintains better alignment with human preferences compared to CLIPScore across all evaluation aspects.}
\label{tab:llava_accuracies}
\end{table*}

This robustness can be attributed to the fact that a weaker text extraction model tends to introduce similar levels of errors across all generated images being compared, thus not significantly affecting their relative rankings in the evaluation. This flexibility allows users to employ any state-of-the-art text extraction model based on their specific requirements and constraints, while maintaining reliable evaluation results.

\section{Evaluation of Image Generation Models using \ours}

\begin{table*}[ht!]
\centering
\rowcolors{2}{gray!25}{white} %
\begin{tabular}{lHHHHHHccc}
    \toprule
    \rowcolor{white} \textbf{Tested Model \quad \quad  \quad} & \textbf{BLEU} & \textbf{BLEU-char} & \textbf{NED} & \textbf{NLCS} & \textbf{Smith Waterman} & \textbf{CLIPScore}($\uparrow$) &  \textbf{\ours}($\uparrow$) & \textbf{Style Fidelity}($\uparrow$) & \textbf{Overall Preference}($\uparrow$) \\
    \midrule
    Stable Diffusion XL & 0.049 $\pm${0.007} & 0.251 $\pm${0.019} & 0.086 $\pm${0.010} & 0.352 $\pm${0.022} & 0.294 $\pm${0.015} & 19.06 $\pm${0.57} & 0.238 $\pm${0.013} & 0.25 & 0.02 \\
    DALL-E 3            & 0.441 $\pm${0.021}  & 0.769 $\pm${0.019}    & 0.673 $\pm${0.024} & 0.807 $\pm${0.015} & 0.776 $\pm${0.017} & 27.25 $\pm${0.39} & 0.739 $\pm${0.018} & \textbf{0.87} & 0.54  \\
    {Stable Diffusion 3} & 0.632 $\pm${0.021}  & 0.836 $\pm${0.017}    & 0.742 $\pm${0.022} & 0.863 $\pm${0.013} &  0.844 $\pm${0.016} & $28.4 \pm 3.7$ & 0.800 $\pm${0.016} & 0.73 & 0.17 \\
    {ideogram}           & \textbf{0.691 $\pm${0.018}} & \textbf{0.878 $\pm${0.010}} & \textbf{0.862 $\pm${0.013}} & \textbf{0.924 $\pm${0.006}} & \textbf{0.899 $\pm${0.009}} & \textbf{29.55 $\pm${0.04}}  & \textbf{0.882 $\pm${0.009}} & \textbf{0.87} & \textbf{0.70}  \\
    \bottomrule
\end{tabular}
\caption{Evaluation of several image generation models using \ours. Ideogram outperforms the others models under \ours. It also garners the top human rating for \textbf{style fidelity} and \textbf{overall preference}. 
}

\label{tab:scoreboard}
\end{table*}

We assess the image generation models mentioned in section~\ref{sec:baseline} using \ours. The results are presented in Table~\ref{tab:scoreboard}. Our evaluation indicates that ideogram attained the highest \ours, with stable diffusion 3 coming in second. Despite DALL-E 3's capability to generate high-quality images, it falls short in accurately rendering the text accurately.

\section{Limitations}
Despite the promising results of \ours in evaluating text fidelity in synthetic images, several limitations must be acknowledged. First, the reliance on existing VLMs for text extraction introduces dependency on their performance and limitations. In scenarios where the VLMs themselves exhibit biases or inaccuracies, these will propagate into our evaluation, potentially skewing \ours results. Moreover, the diverse nature of text styles and formats in \oursdata may not comprehensively cover all real-world use cases. Our metric is evaluated on Latin text and should benefit from being evaluated with non-Latin text as well. Lastly, our human evaluation process, although extensive, is subject to individual annotator biases and interpretations. While we have taken measures to ensure consistency and reliability, human evaluations inherently carry a degree of subjectivity that can influence the assessment outcomes.

\section{Conclusion}
We introduced a comprehensive evaluation framework, \ours, designed to assess the fidelity of text embedded in synthetic images generated by various models. Our framework evaluates the degree to which the generated images accurately follow textual instructions, leveraging a combination of automatic metrics based on human judgment. By comparing the performance of our metrics with human preferences, we demonstrated that \ours aligns more closely with human judgment compared to traditional metrics like CLIPScore, indicating its efficacy for evaluating text fidelity in image generation models. In future work, we aim to explore the possibility of extending our approach to evaluate image generation in general domains beyond text rendering. We also plan to assess whether calculating the likelihood $q_\phi(t|i)$ could provide a more precise evaluation of text fidelity compared to the dissimilarity metrics we used in \ours.

\section{Acknowledgement}
\input{sec/ask}

\bibliography{references}
\bibliographystyle{plain}

\newpage

\appendix

\input{sec/app}

\end{document}

%% file: sec/ask.tex
We would like to thank Ziv Wolkowicki, Barry Theobald, Samy Bengio, Richard Bai, Zijin Gu, and Tatiana Likhomanenko for their valuable feedback and contributions.

%% file: sec/app.tex
\onecolumn
\begin{center}
    {\Large \bf Appendix}
\end{center}

\section{Proof of \eqref{eq:mi}}
\label{app:proof}

\begin{align}
    \mathbf{MI}_\theta(i, t) &= \mathbb{E}_{t \sim \mathcal{T}, i \sim p_\theta(\cdot|t)} \log \frac{p(i, t)}{p(i) p(t)} \nonumber \\ 
    & = H(t) + \mathbb{E}_{i} D_{KL} (p_\theta(t|i), q_\phi(t|i)) + \mathbb{E}_{t \sim \mathcal{T}, i \sim p_\theta(\cdot|t)} \log q_\phi(t|i) \nonumber \\
    &\geq \mathbb{E}_{t \sim \mathcal{T}} \mathbb{E}_{i \sim p_\theta(\cdot|t)} \log q_\phi(t|i) \triangleq \mathcal{L}_{\mathbf{MI}}(\theta;\phi). \nonumber
\end{align}

where $H(\cdot)$ represents the entropy. $D_{KL}$ denotes the KL divergence between two distributions. 

\section{Dataset composition}
\label{dataset_composition}

We provide the dataset composition of \oursdata below. \ours covers a variety of scenarios that instruct the image generation model to render text. 

\begin{figure}[h]
\centering
\includegraphics[width=0.99\textwidth]{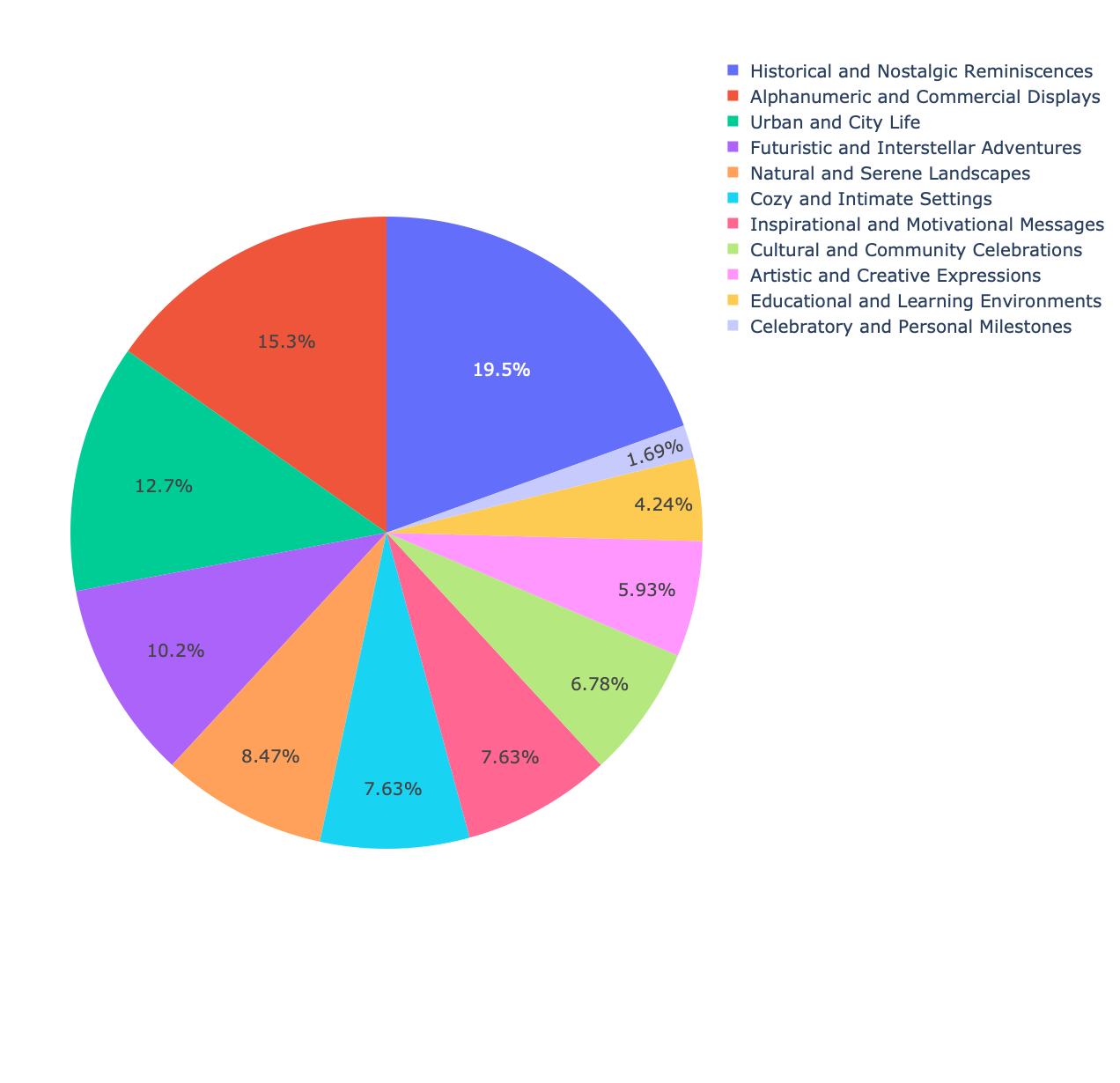}
\vspace{-20mm}
\caption{Composition of \oursdata dataset.}
\label{fig:lost_in_the_middle}
\end{figure}
\clearpage

\section{Image description methods}

\subsection{\textbf{OCR+GPT-4o} prompt}
\label{gpt4o_ocr_prompt}

When combining \textbf{OCR+GPT-4o}, we used the prompt below to generate a response. In the prompt. \textcolor{blue}{\{ocr\_extracted\_caption\}} refers to the caption extracted by OCR. 

\begin{center}
\begin{tcolorbox}[width=0.99\textwidth]
\textit{This image contains a main quote and it might contain additional text. We already extracted both the main quote and any additional text from the image, and it follows: \textcolor{blue}{\{ocr\_extracted\_caption\}}. We want to isolate only the main quote. From this text, identify the main quote and extract it in the right order, without correcting any typos or issues you may encounter, and without adding any new words. Output the main quote between quotes and do not output anything else. 
}
\end{tcolorbox}
\end{center}

\subsection{GPT-4o versus LLaVa}

We assess the image generation models mentioned in section~\ref{sec:baseline} using \ours with LLaVA-NeXT. The results are presented in Table~\ref{tab:scoreboard_llava}. As an open-source model, LLaVa offers lower costs compared to GPT-4o, making it a cost-effective alternative for \ours.

\begin{table*}[ht!]
\centering
\rowcolors{2}{gray!25}{white} %
\begin{tabular}{lcc}
    \toprule
    \rowcolor{white} \textbf{Tested Model \quad \quad  \quad} & \textbf{\ours (GPT-4o)} & \textbf{\ours (LLaVa)} \\
    \midrule
    Stable Diffusion XL & 0.238 $\pm${0.013} & 0.357 $\pm${0.088} \\
    DALL-E 3            & 0.739 $\pm${0.018} & 0.746 $\pm${0.018} \\
    {Stable Diffusion 3} & 0.800 $\pm${0.016} & 0.816 $\pm${0.017} \\
    {ideogram}           & \textbf{0.882 $\pm${0.009}} & \textbf{0.855 $\pm${0.013}} \\
    \bottomrule
\end{tabular}
\caption{Evaluation of several image generation models using \ours with LLaVa. Similarly to \ours with GPT-4o, ideogram outperforms the other models under \ours with LLaVa.} 
\label{tab:scoreboard_llava}
\end{table*}

\section{Human evaluation details}
\label{app:human_eval}

In total, we recruited 104 human annotators to participate in the study using our internal crowd-source annotation platform. Annotators were recruited from Canada, Great Britain, the United States, Australia, Singapore and India and paid an average of \$76.28 USD per hour.

To ensure the quality of the annotation data, we qualified each rater by assigning a set of test questions for which the answers were known. In order to be qualified to annotate the study, raters were required to correctly answer at least 90\% of the test set questions. Additionally, we manually inspected the annotations to validate the human ratings. We found that the inter-rater agreement was high, with only 4.87\% of the tasks requiring two additional judges to reach an agreement. 

\begin{figure*}[h!]
 \vspace{-8mm}
 \includegraphics[width=0.99\textwidth, trim={0 8mm 0 0},clip]{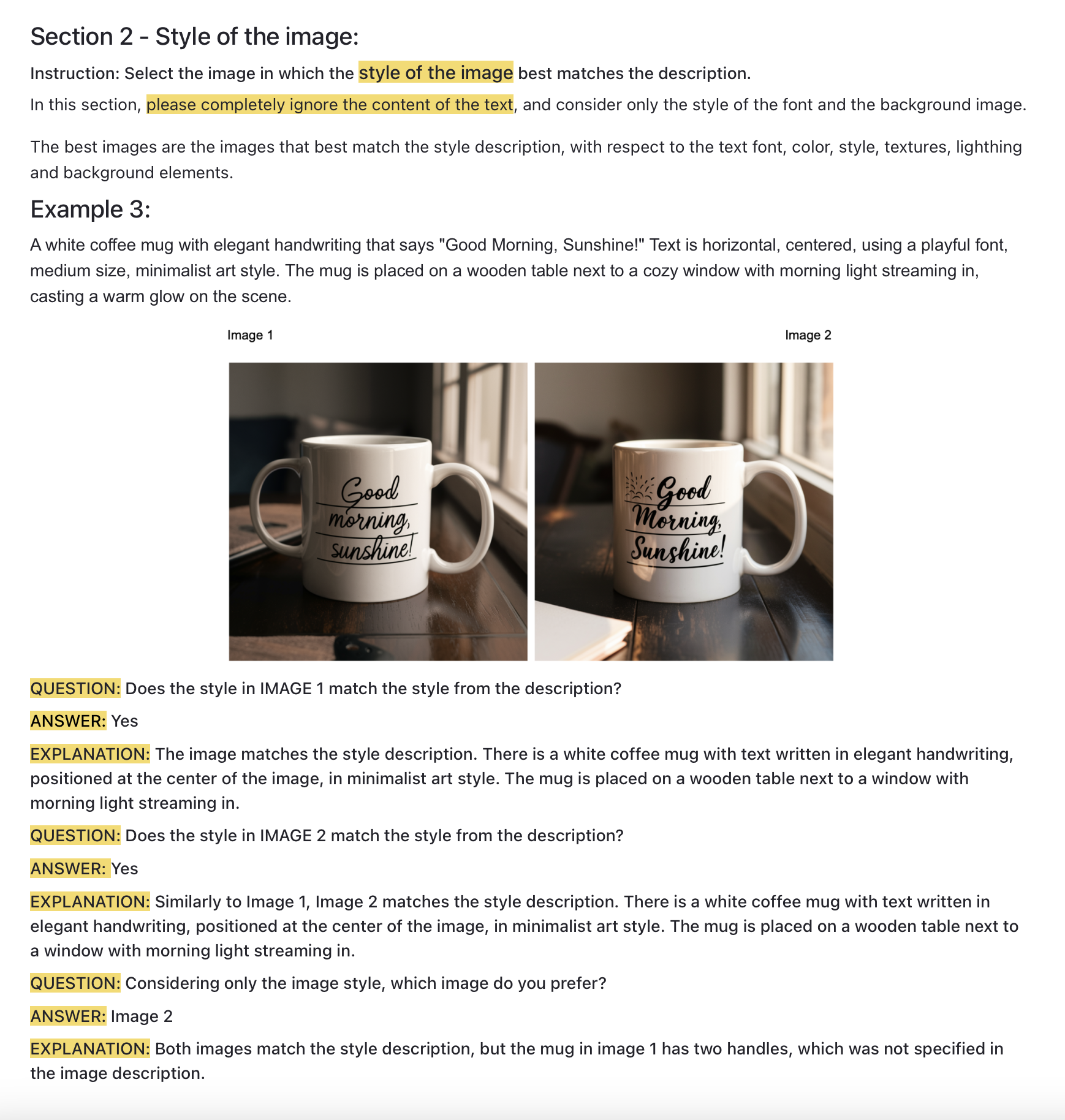} 
 \includegraphics[width=0.99\textwidth, trim={0 1mm 0 0},clip]{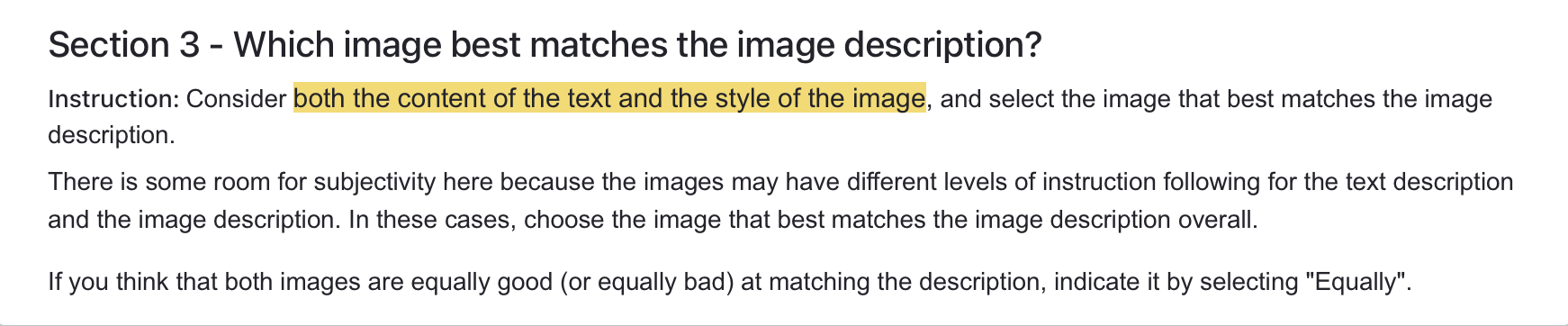} 
 \caption{The tutorial UI is split into 3 sections, corresponding to the sections of the annotation task: text fidelity, style fidelity, and overall preference. Each section contains example image pairs that demonstrate potential issues annotators might encounter, along with the correct answers for each scenario. }
\label{app:guidelines}
\end{figure*}

\begin{figure*}[h!]
 \includegraphics[width=0.99\textwidth, trim={0 0 0 0},clip]{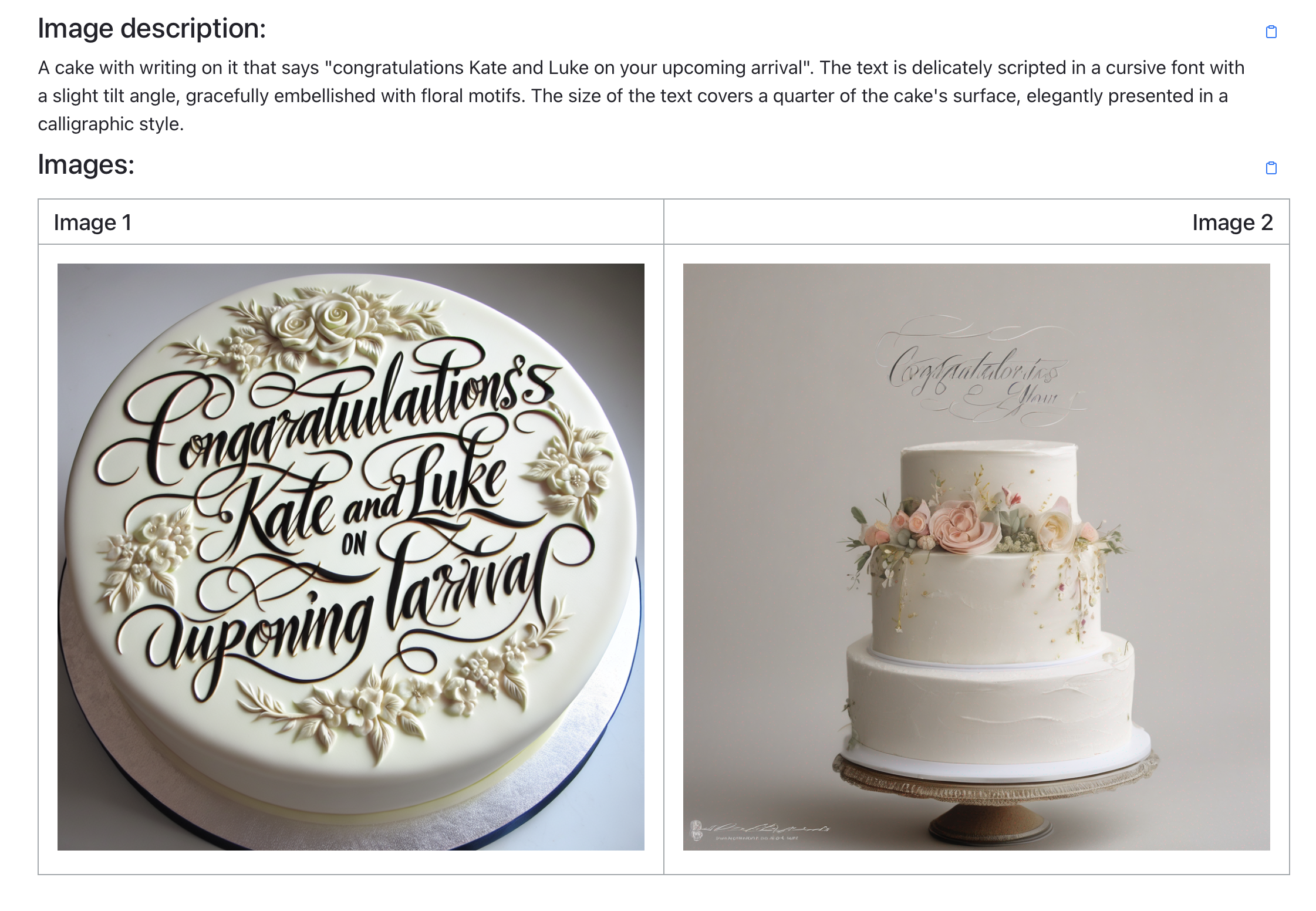} 
 \includegraphics[width=0.99\textwidth, trim={0 6mm 0 0},clip]{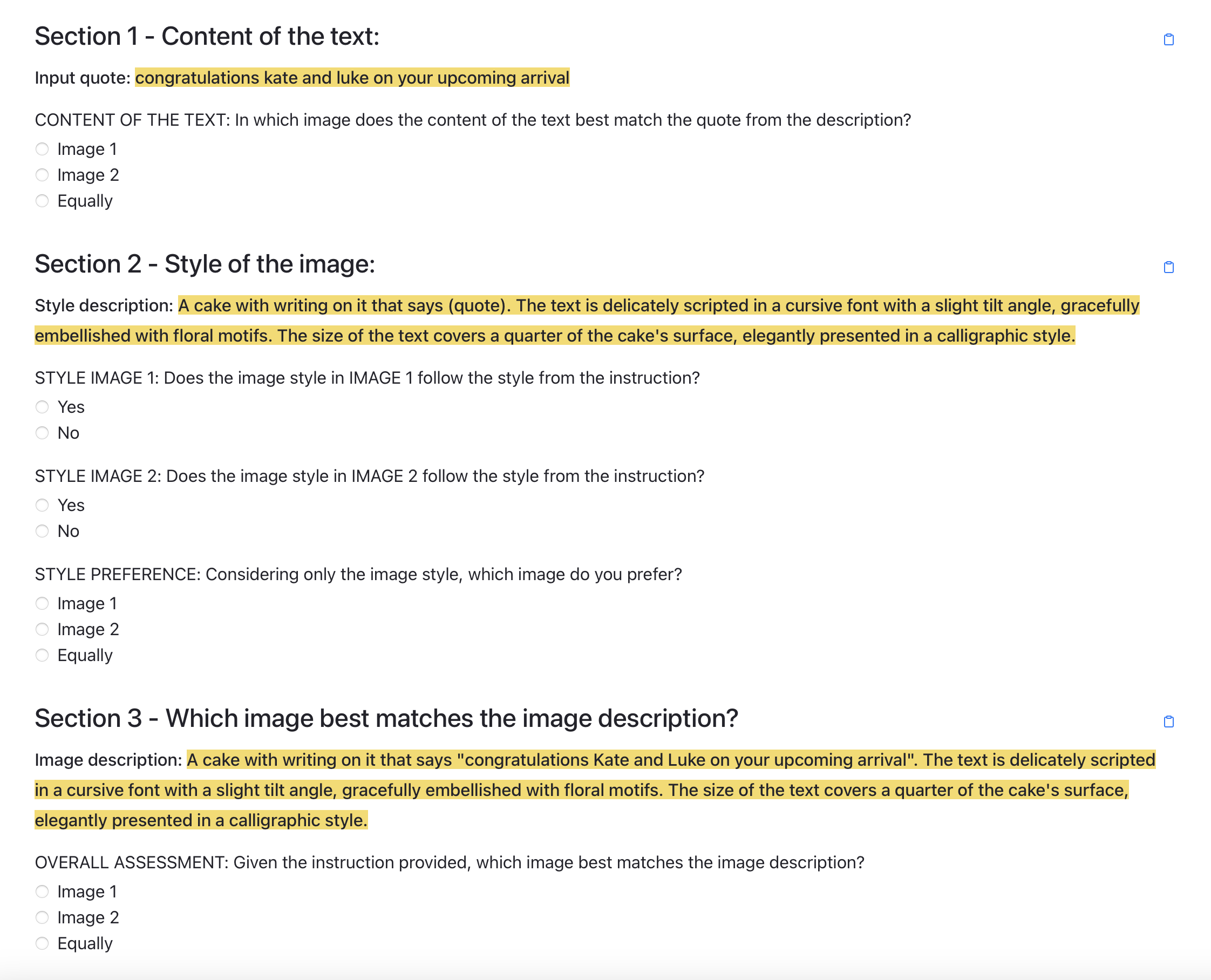} 
\caption{Annotation task UI. Users we provided with an image description and a pair of images, and asked to rate the images with respect to their text fidelity, style fidelity, and overall preference.}
\label{app:annotation_task}
\end{figure*}

We show the UI for the annotation task in Figures \ref{app:guidelines} and \ref{app:annotation_task}: Figure \ref{app:guidelines} shows the UI for the tutorial that was provided to each annotator, and Figure \ref{app:annotation_task} shows the task UI. 

\subsection{Tutorial UI}
\begin{figure*}[h!]
 \vspace{-8mm}
 \includegraphics[width=0.99\textwidth, trim={0 0 0 1mm},clip]{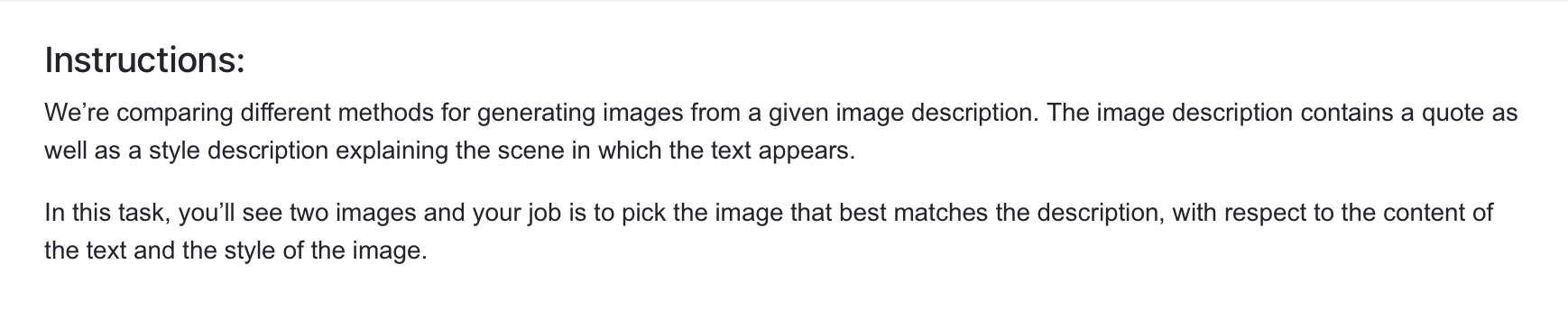} 
 \includegraphics[width=0.99\textwidth]{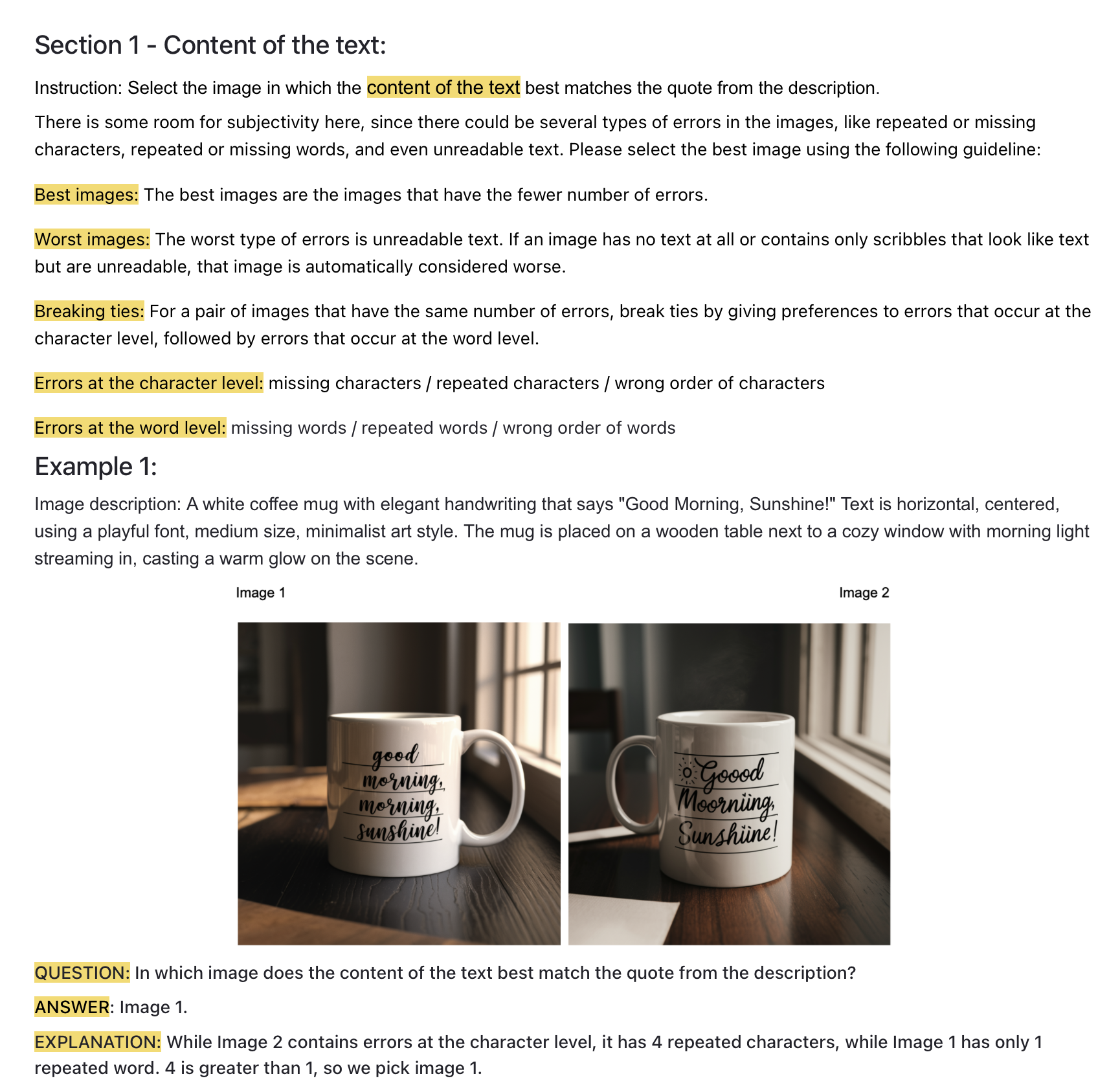} 
  \includegraphics[width=0.99\textwidth]{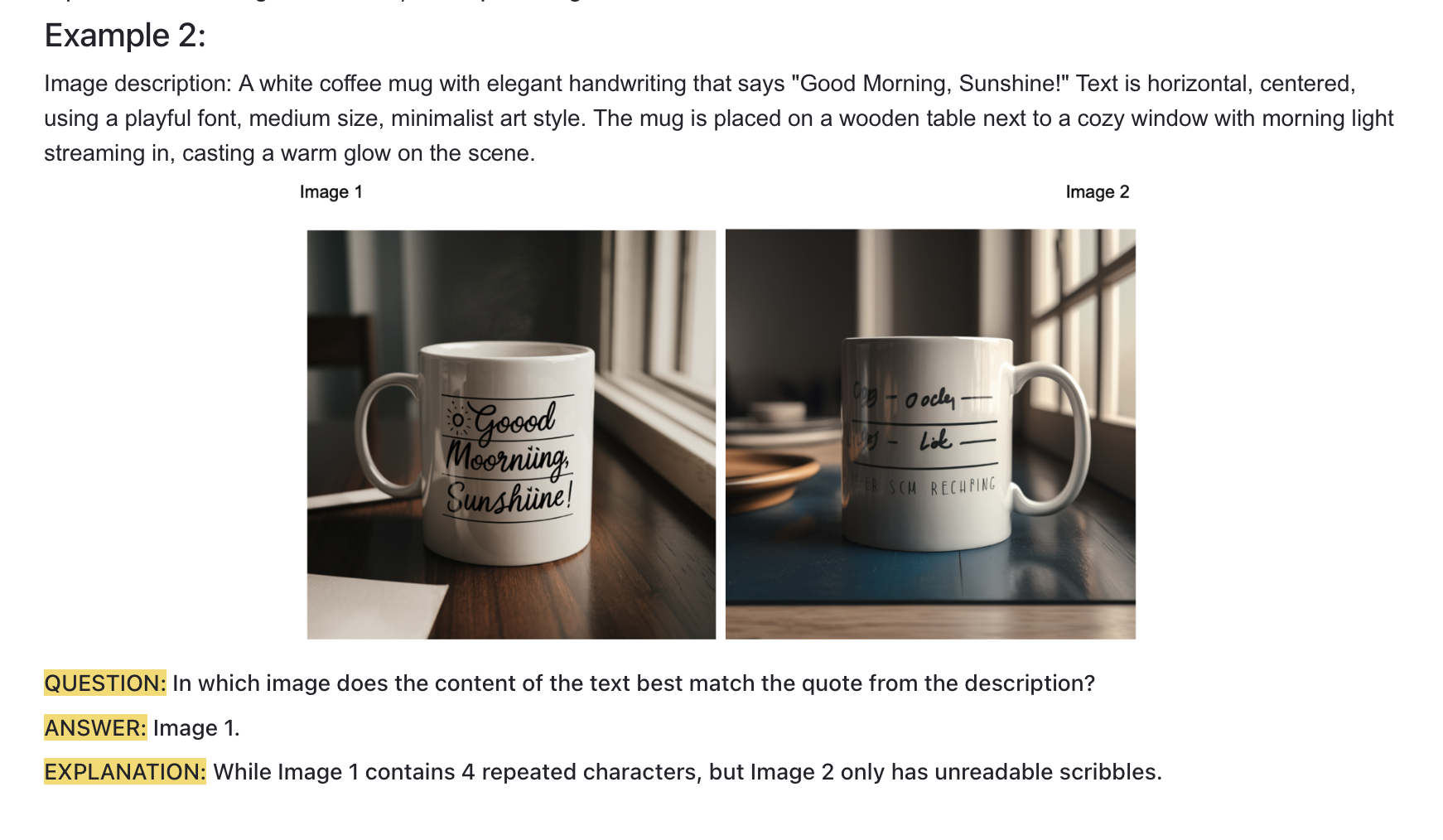} 
 \vspace{5mm}
\end{figure*}

\clearpage

\subsection{Annotation Task UI}
\vspace{1mm}
\clearpage

\section{\ours Sensitivities}

\subsection{Sensitivity to the length of the instruction}
\label{text_length}

Table~\ref{tab:text_length_correlations} shows the Pearson correlation coefficients between different variants of \ours and overall input caption lengths across different image generation models. This suggests that \ours exhibits minimal correlation with caption length for each model, indicating it yields stable score across various lengths of the input text.

\begin{table*}[ht!]
\centering
\rowcolors{2}{gray!25}{white} %
\begin{tabular}{lcccc}
    \toprule
    \rowcolor{white} \textbf{\ours variants:} & SD XL & DALL-E 3 & SD 3 & ideogram\\
    \midrule
    {\ours (NED)} & 0.00 &  0.01 &  0.00 & 0.00 \\
    {\ours (BLEU)} & 0.06 &  0.03 &  0.12 & 0.04 \\
    {\ours (character-BLEU)} & 0.03 & 0.04 & 0.02 & 0.07\\
    {\ours (NLCS)} & 0.04 & 0.04 & 0.02 & 0.05 \\
    {\ours (Smith Waterman)} & 0.10 &  0.02 & 0.03 & 0.03\\
    {\ours (Ensemble Score)} & 0.03 & 0.01 & 0.02 & 0.02\\
    \bottomrule
\end{tabular}
\caption{Pearson correlation coefficients between \ours and different input caption lengths. SD denotes Stable Diffusion. The results show no significant correlation across various image generation models, demonstrating that \ours remains stable across various text lengths.}
\label{tab:text_length_correlations}
\end{table*}

\subsection{Sensitivity to text recaptioning}
\label{recaptioning}

\begin{figure}[h]
\centering
\includegraphics[width=0.8\textwidth]{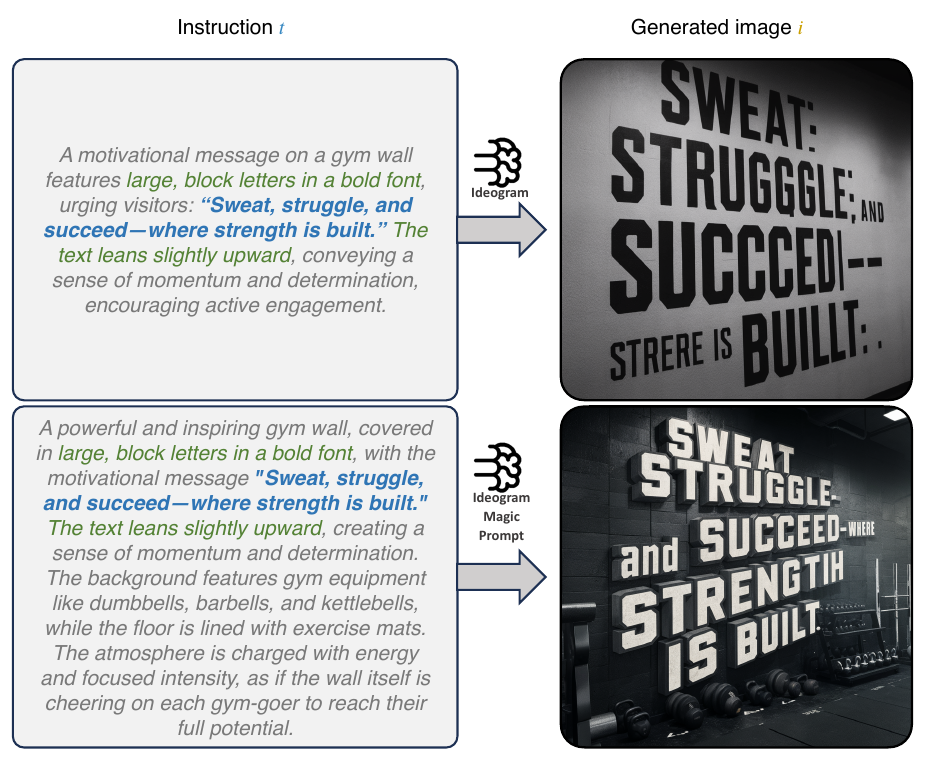}
\caption{The ideogram Magic Prompt model augments the \textcolor{gray}{contextual information} of the input instruction, adding more stylistic details. We found that this helps control the text fidelity evaluation of the rendered text.}
\label{fig:recaptioning_qualitative}
\end{figure}

We found that augmenting the input instruction with more detailed stylistic details and contextual information helps control the text fidelity evaluation of the rendered text. The results are shown in Table~\ref{tab:recaptioning}. We assessed this by comparing the quantitative results of ideogram and ideogram Magic Prompt \ref{recaptioning}. The ideogram Magic Prompt model is an extension of the default ideogram model, where an augmented image instruction is suggested and used to generate the image \ref{fig:recaptioning_qualitative}. 

It can be observed that \ours has a higher mean value and slightly lower variance, demonstrating that incorporating richer stylistic nuances contributes to a more controlled setting for text fidelity evaluation. 

\begin{table*}[ht!]
\centering
\rowcolors{2}{gray!25}{white} %
\begin{tabular}{lcc}
    \toprule
    \rowcolor{white} \textbf{\ours variants:} & ideogram & ideogram Magic Prompt\\
    \midrule
    {\ours (NED) ($\downarrow$)} & 0.138 $\pm{0.013}$ & \textbf{0.124 $\pm${0.014}}\\
    {\ours (BLEU) ($\uparrow$)} & 0.691 $\pm{0.018}$ & \textbf{0.772 $\pm${0.015}}\\
    {\ours (character-BLEU) ($\uparrow$)} & 0.878 $\pm{0.010}$ & \textbf{0.912 $\pm${0.009}}\\
    {\ours (NLCS) ($\uparrow$)} & 0.924 $\pm{0.006}$ & \textbf{0.936 $\pm${0.006}}\\
    {\ours (Smith Waterman) ($\uparrow$)} & 0.899 $\pm{0.009}$ & \textbf{0.909 $\pm${0.009}}\\
    \midrule
    {\ours (Ensemble Score) ($\uparrow$)} & 0.882 $\pm{0.009}$ & \textbf{0.895 $\pm${0.009}}\\
    \bottomrule
\end{tabular}
\caption{ideogram with recaptioning consistently outperforms ideogram with standard prompts.}
\label{tab:recaptioning}
\end{table*}

\section{Comparative qualitative analysis via example generations}
\label{text_fidelity}

In the following sections, we present example generations from each image generation model. Each model's outputs are divided into two columns: left and right. The left column showcases the most faithful generations from the model, while the right column displays examples with the lowest text fidelity. These low-fidelity examples often feature numerous typos, repeated words and characters, illegible glyphs, or a complete absence of text. These comparisons help elucidate the differences in text fidelity among the models.
\clearpage

\subsection{Stable Diffusion XL}

\begin{table*}[ht!]
\centering
\vspace{-2mm}
\begin{tabular}{ll}
\includegraphics[width=.5\linewidth,valign=m,height=.2\linewidth]{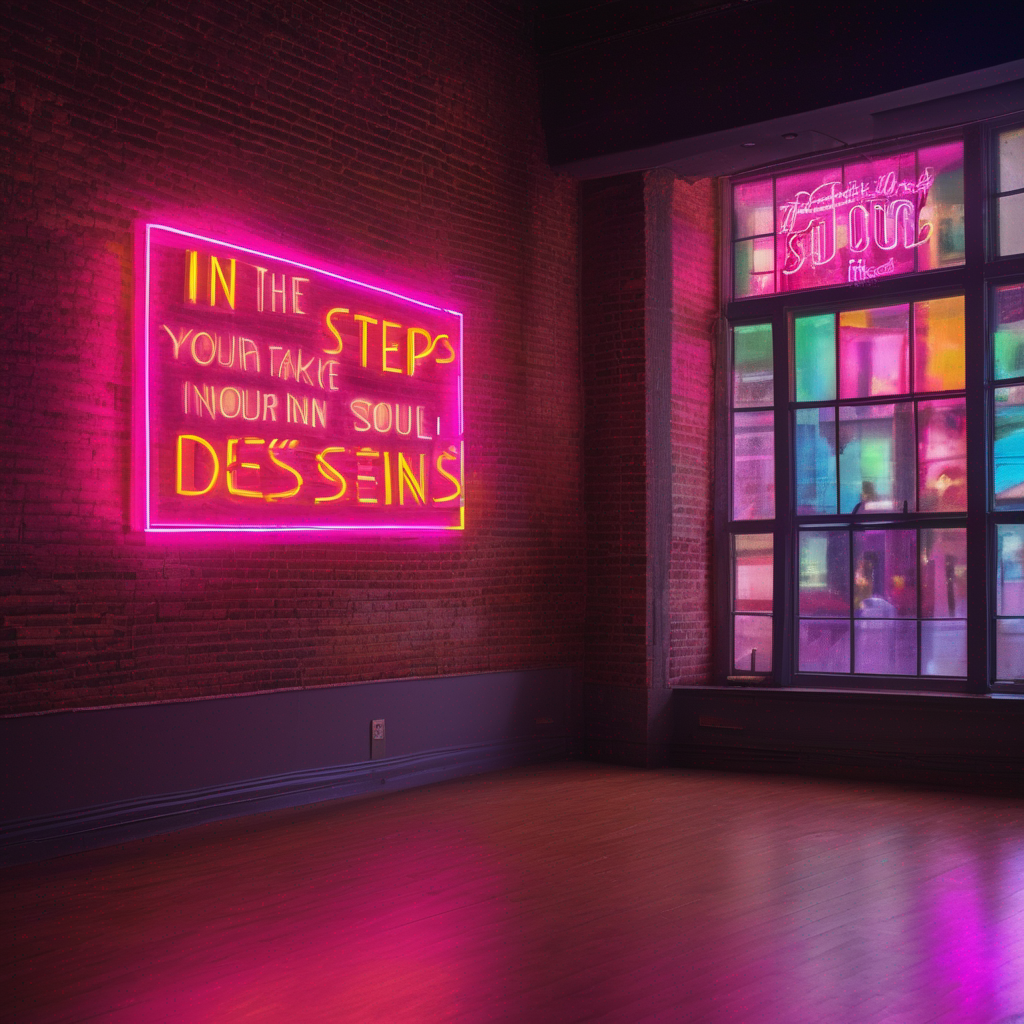} & \includegraphics[width=.5\linewidth,valign=m,height=.2\linewidth]{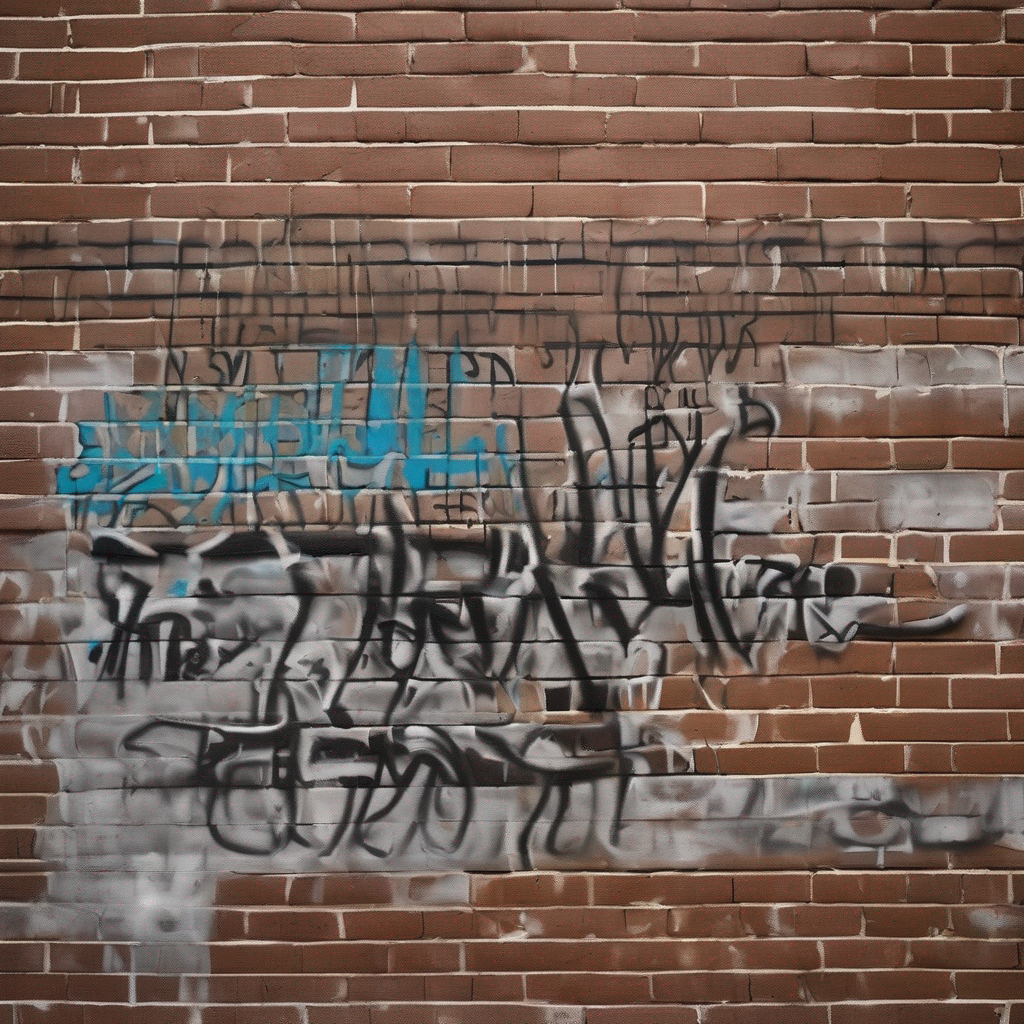} \\ \addlinespace
\includegraphics[width=.5\linewidth,valign=m,height=.2\linewidth]{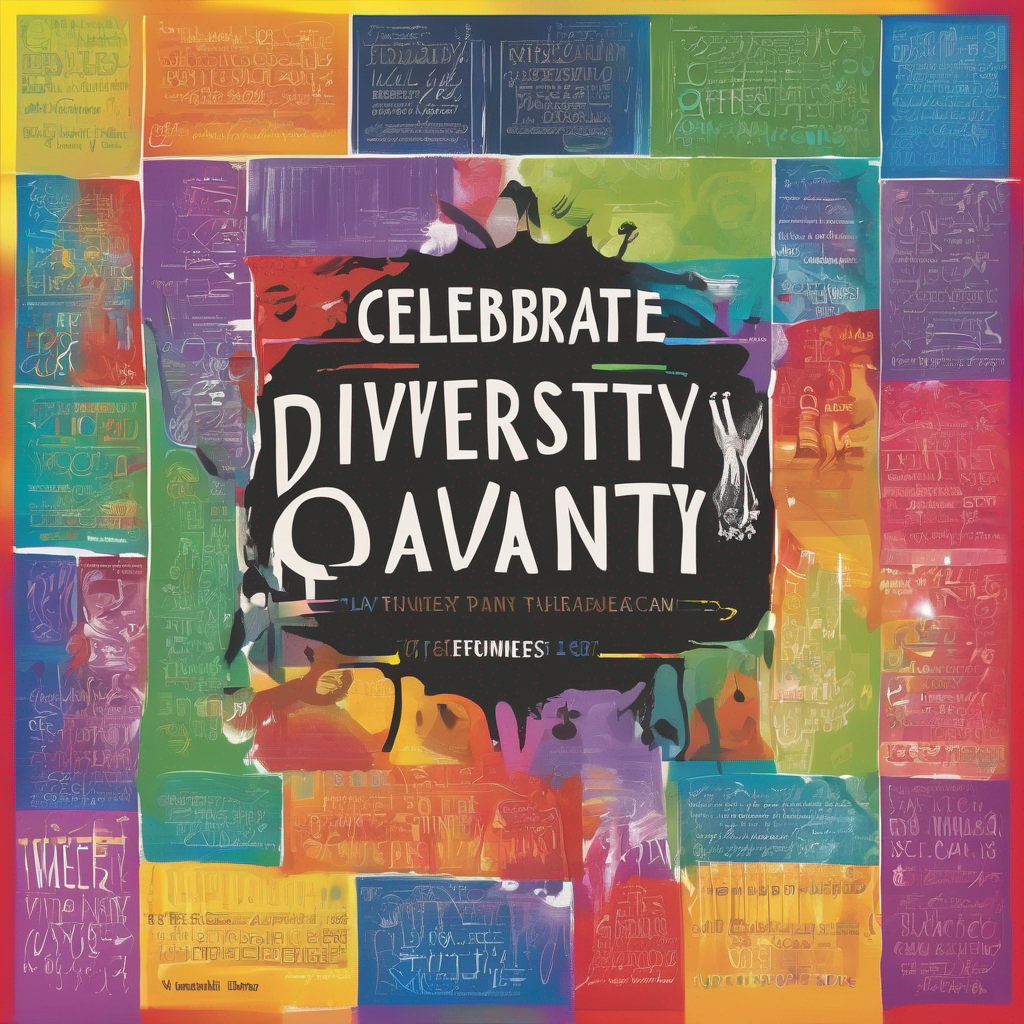} & \includegraphics[width=.5\linewidth,valign=m,height=.2\linewidth]{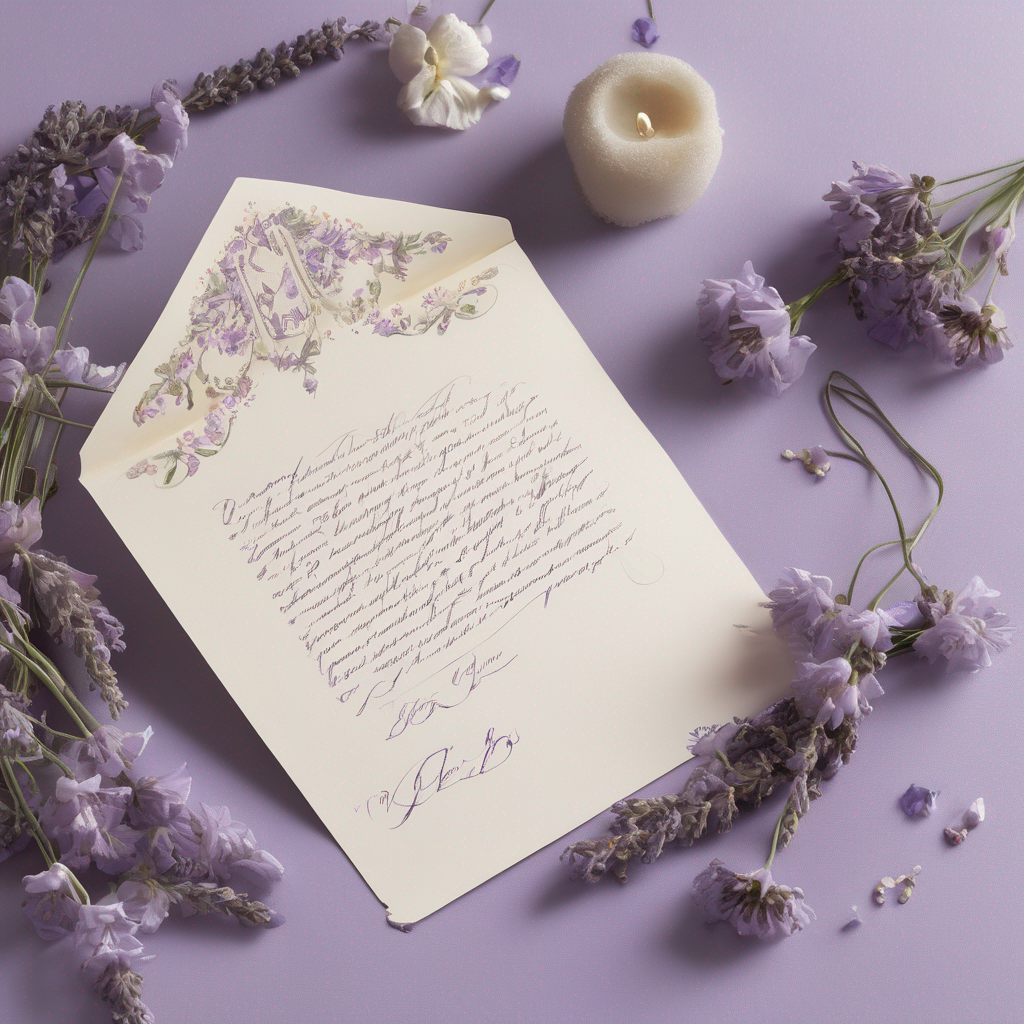} \\ \addlinespace
\includegraphics[width=.5\linewidth,valign=m,height=.2\linewidth]{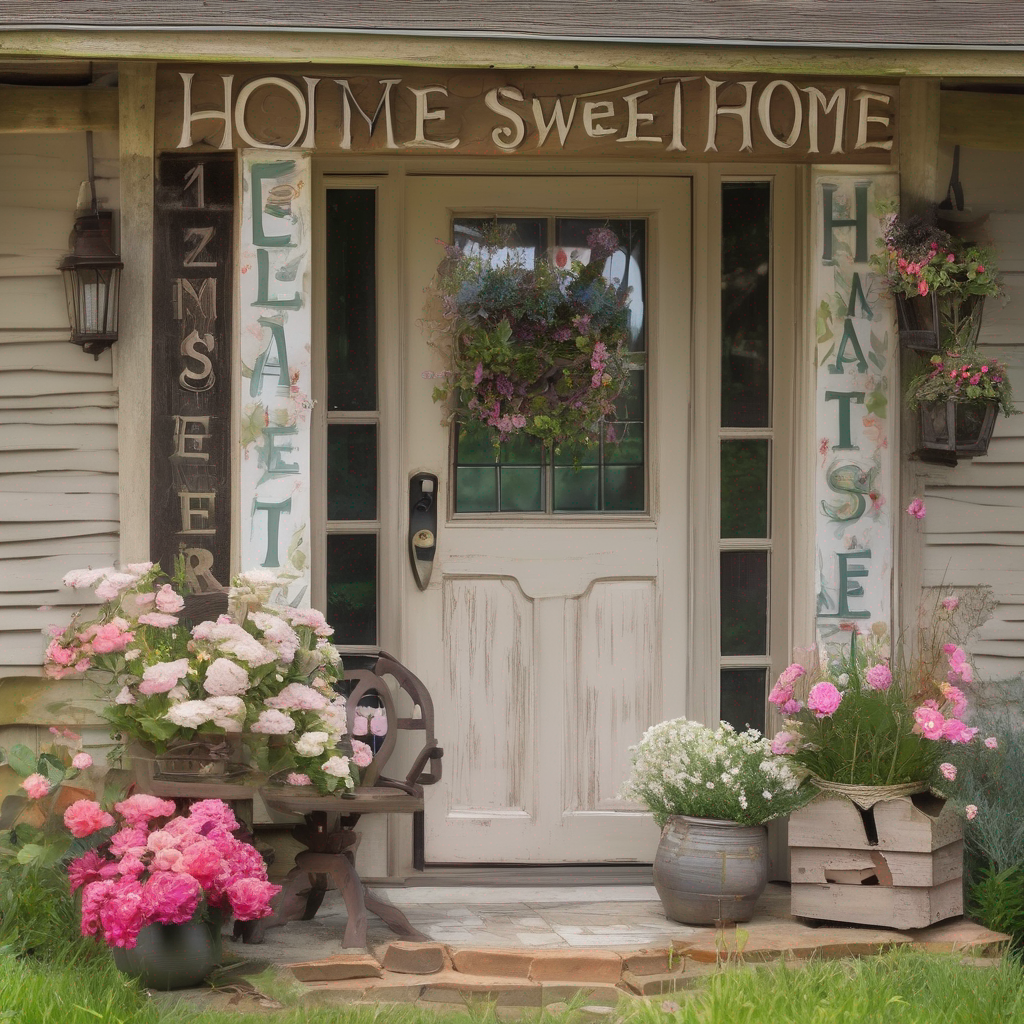} & \includegraphics[width=.5\linewidth,valign=m,height=.2\linewidth]{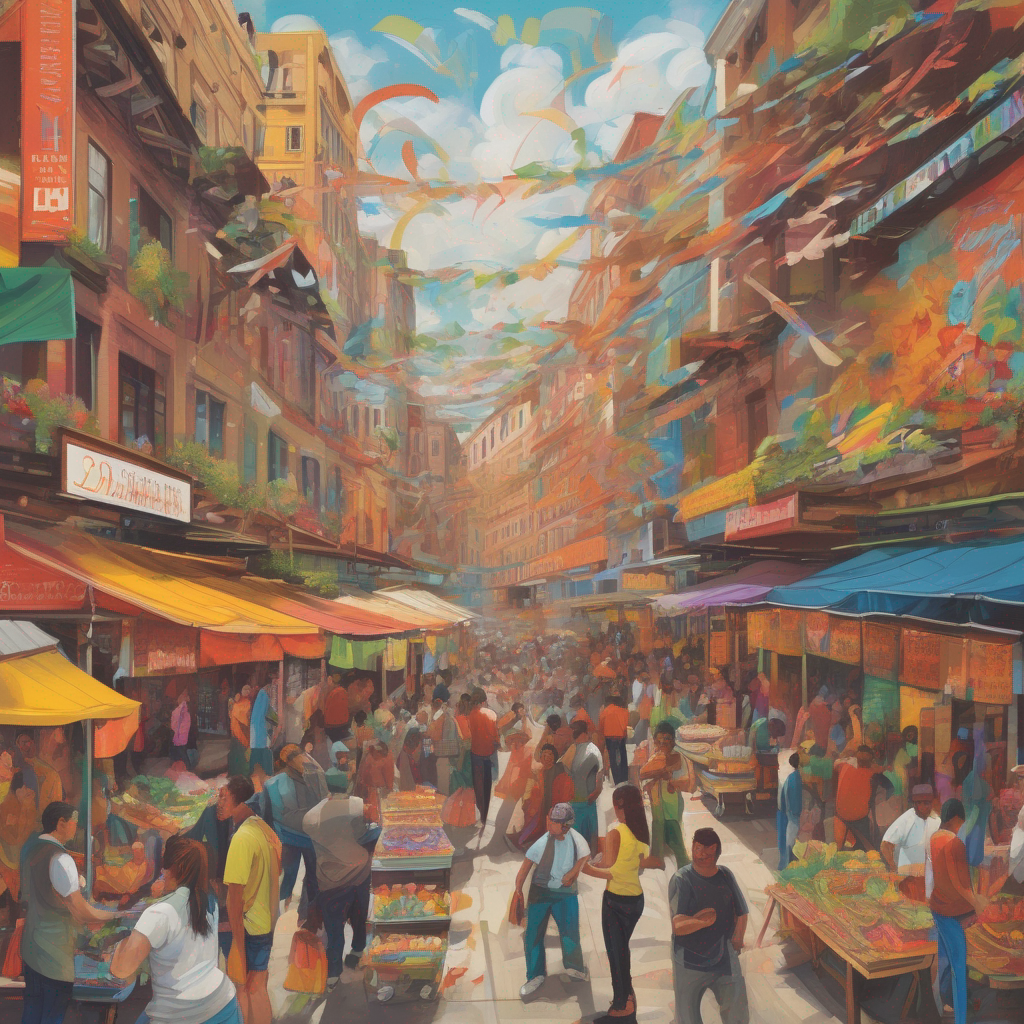} \\ \addlinespace
\includegraphics[width=.5\linewidth,valign=m,height=.2\linewidth]{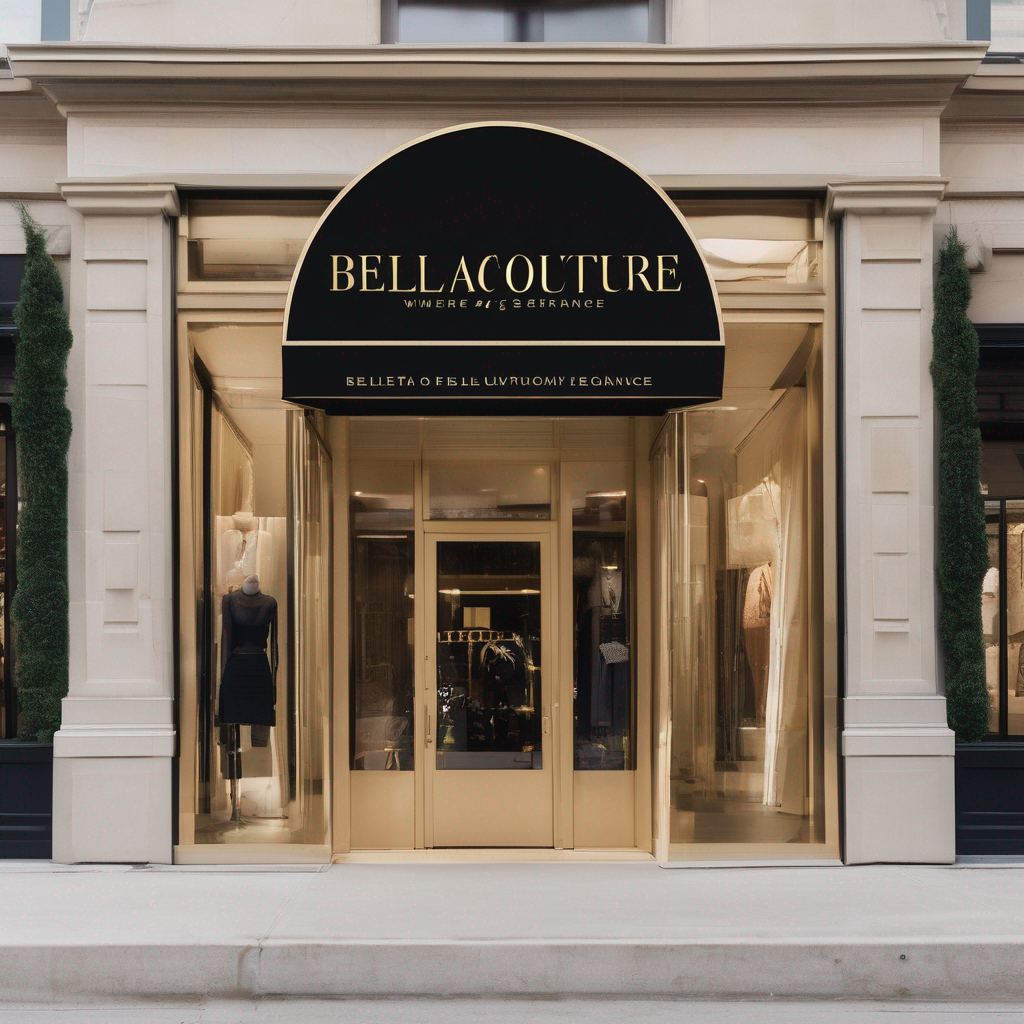} & \includegraphics[width=.5\linewidth,valign=m,height=.2\linewidth]{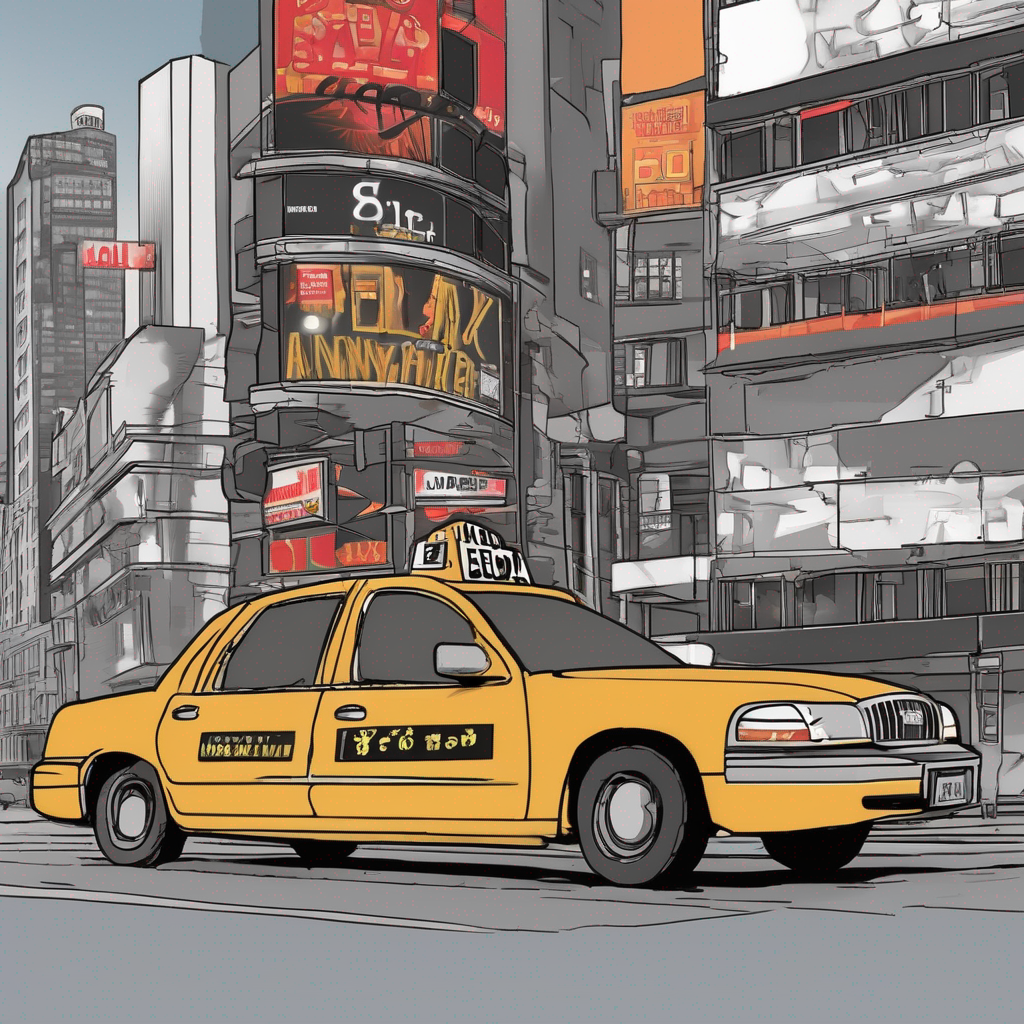} \\ \addlinespace
\end{tabular}
\vspace{-3mm}
\caption{Stable Diffusion XL example generations. On the \textbf{left}, see some generations with higher text fidelity. On the \textbf{right}, see some generations with lower text fidelity.}
\vspace{-20mm}
\end{table*}
\clearpage

\subsection{DALLE 3 }
\begin{table*}[ht!]
\centering
\vspace{-2mm}
\begin{tabular}{ll}
\includegraphics[width=.5\linewidth,valign=m,height=.2\linewidth]{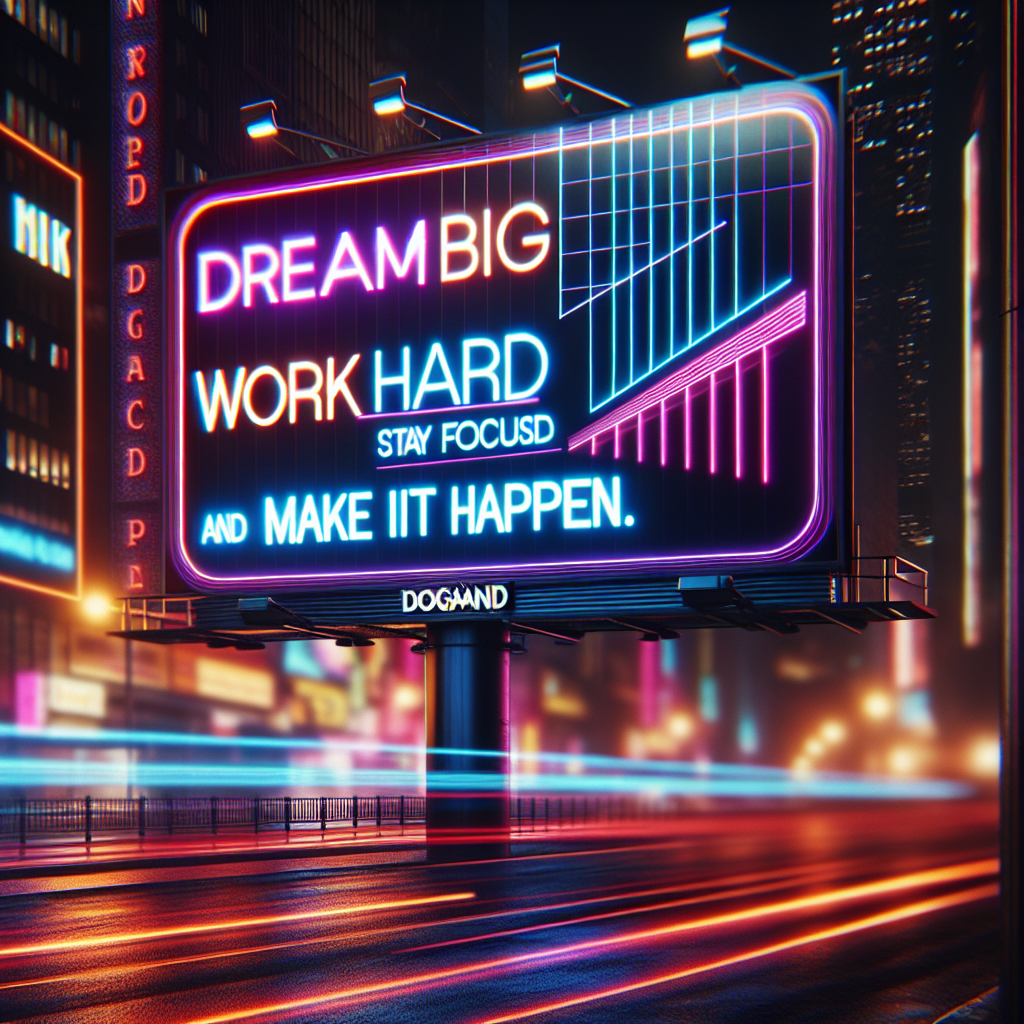} & \includegraphics[width=.5\linewidth,valign=m,height=.2\linewidth]{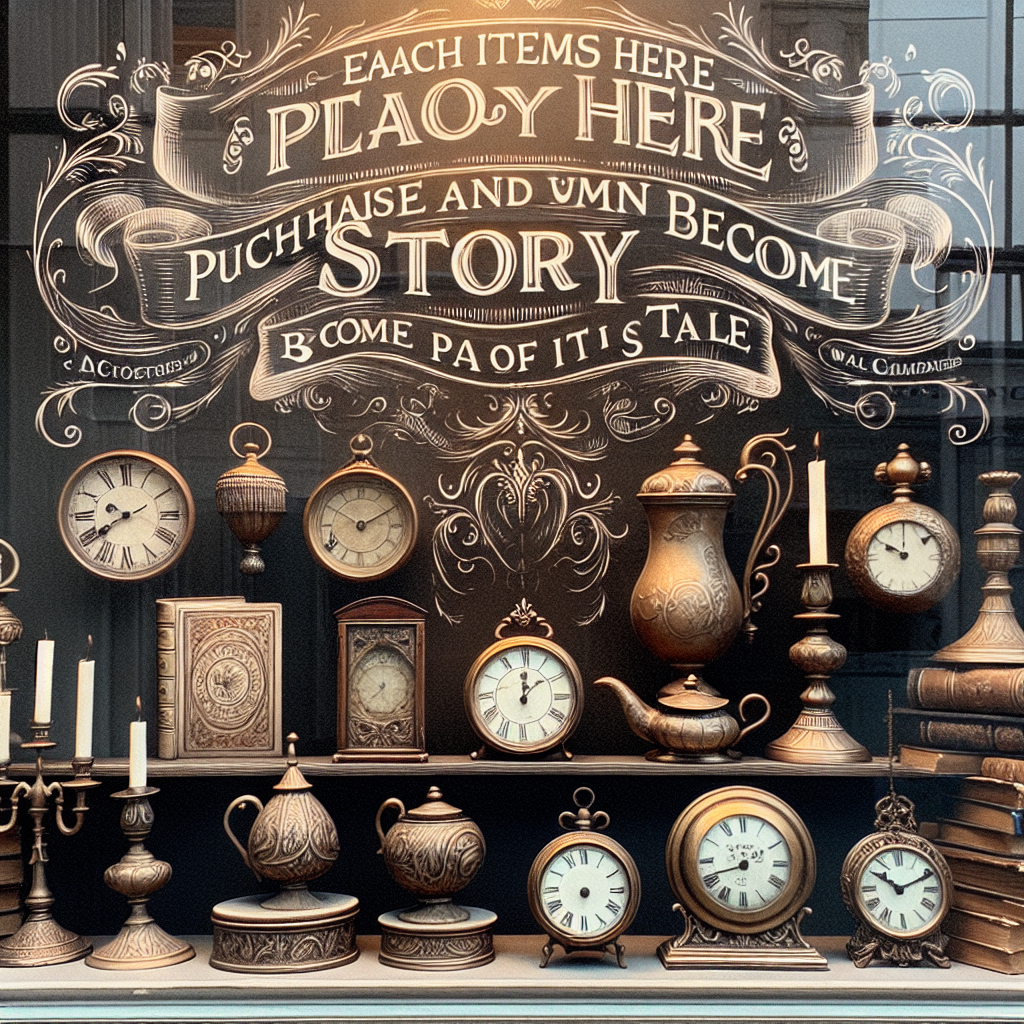} \\ \addlinespace
\includegraphics[width=.5\linewidth,valign=m,height=.2\linewidth]{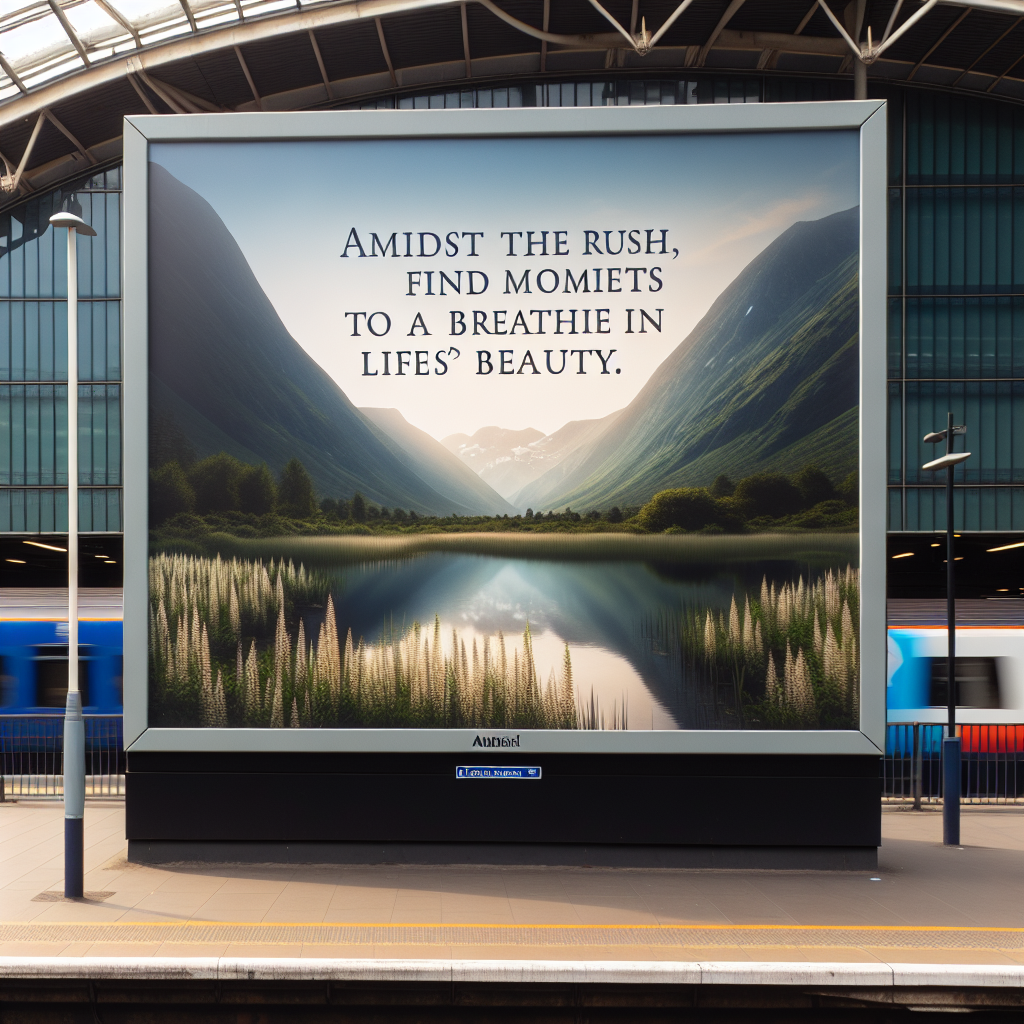} & \includegraphics[width=.5\linewidth,valign=m,height=.2\linewidth]{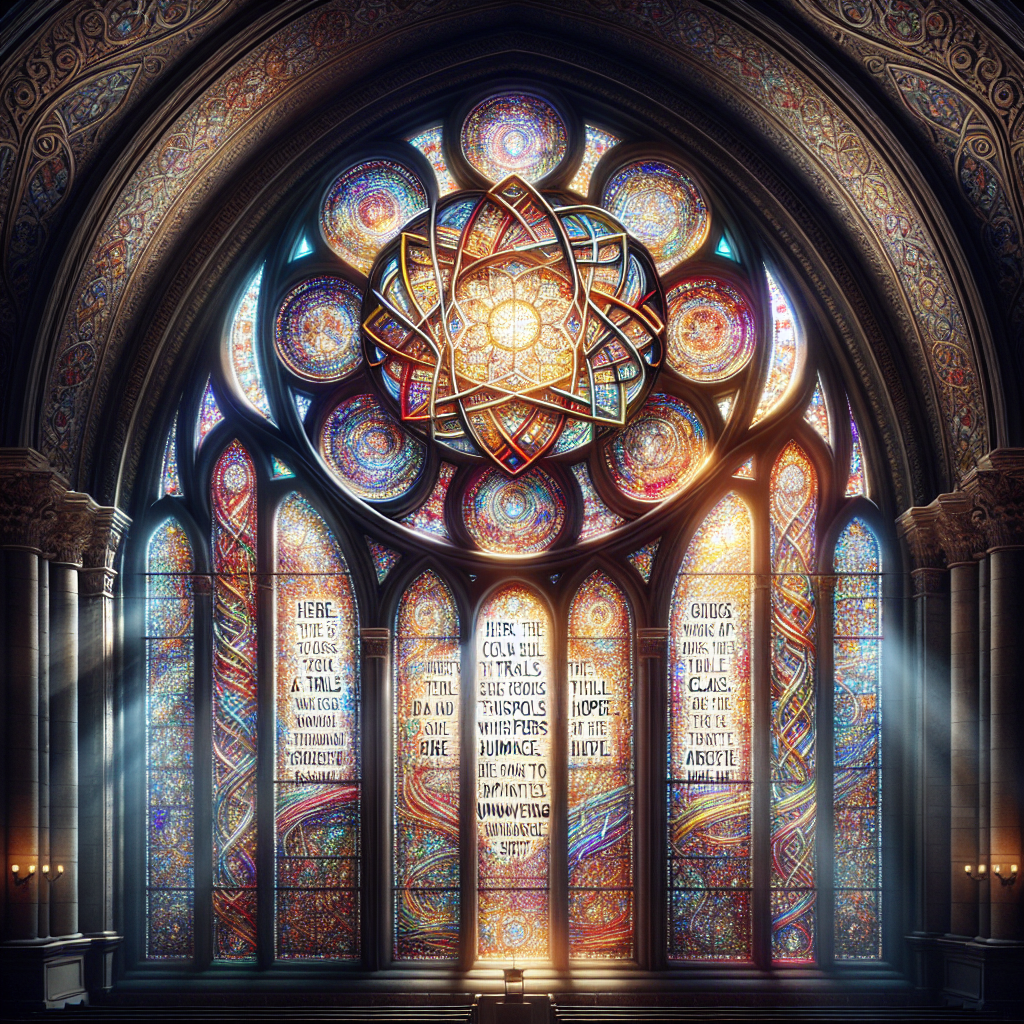} \\ \addlinespace
\includegraphics[width=.5\linewidth,valign=m,height=.2\linewidth]{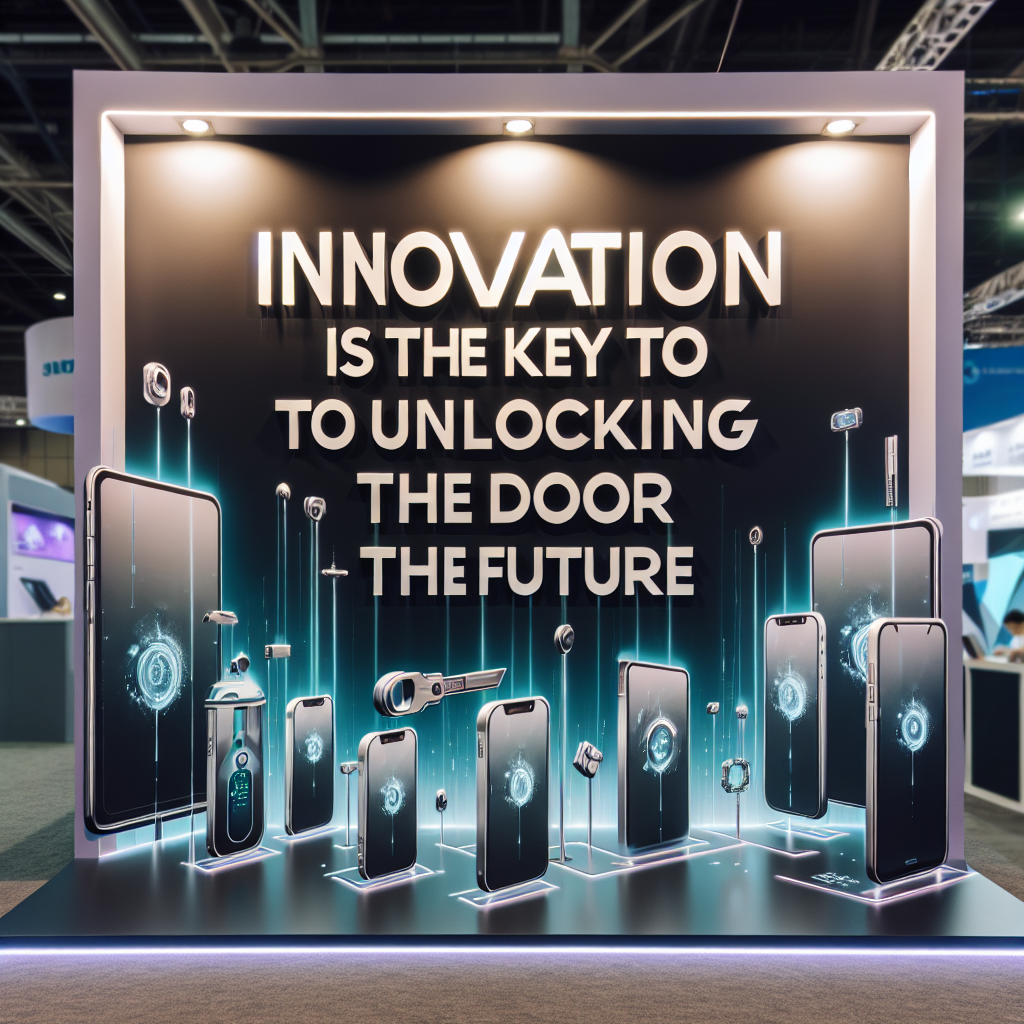} & \includegraphics[width=.5\linewidth,valign=m,height=.2\linewidth]{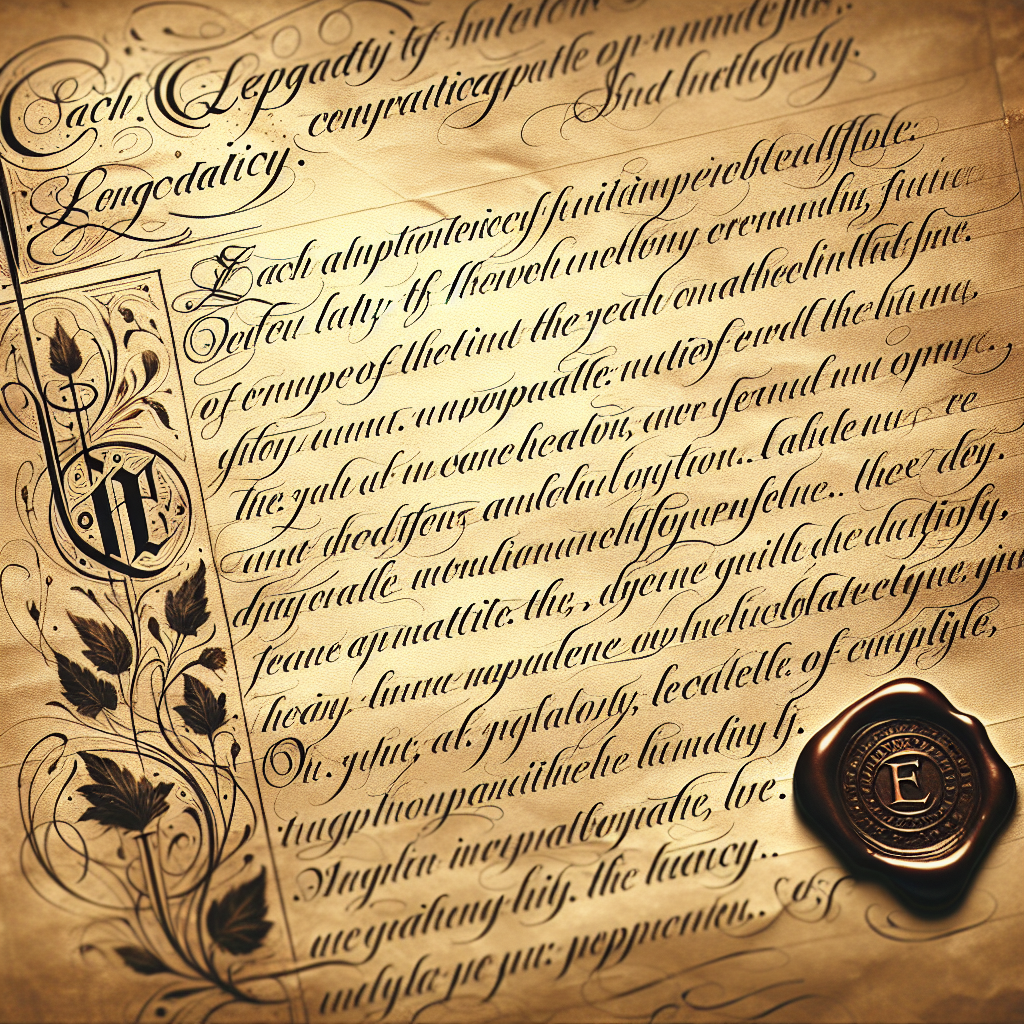} \\ \addlinespace
\includegraphics[width=.5\linewidth,valign=m,height=.2\linewidth]{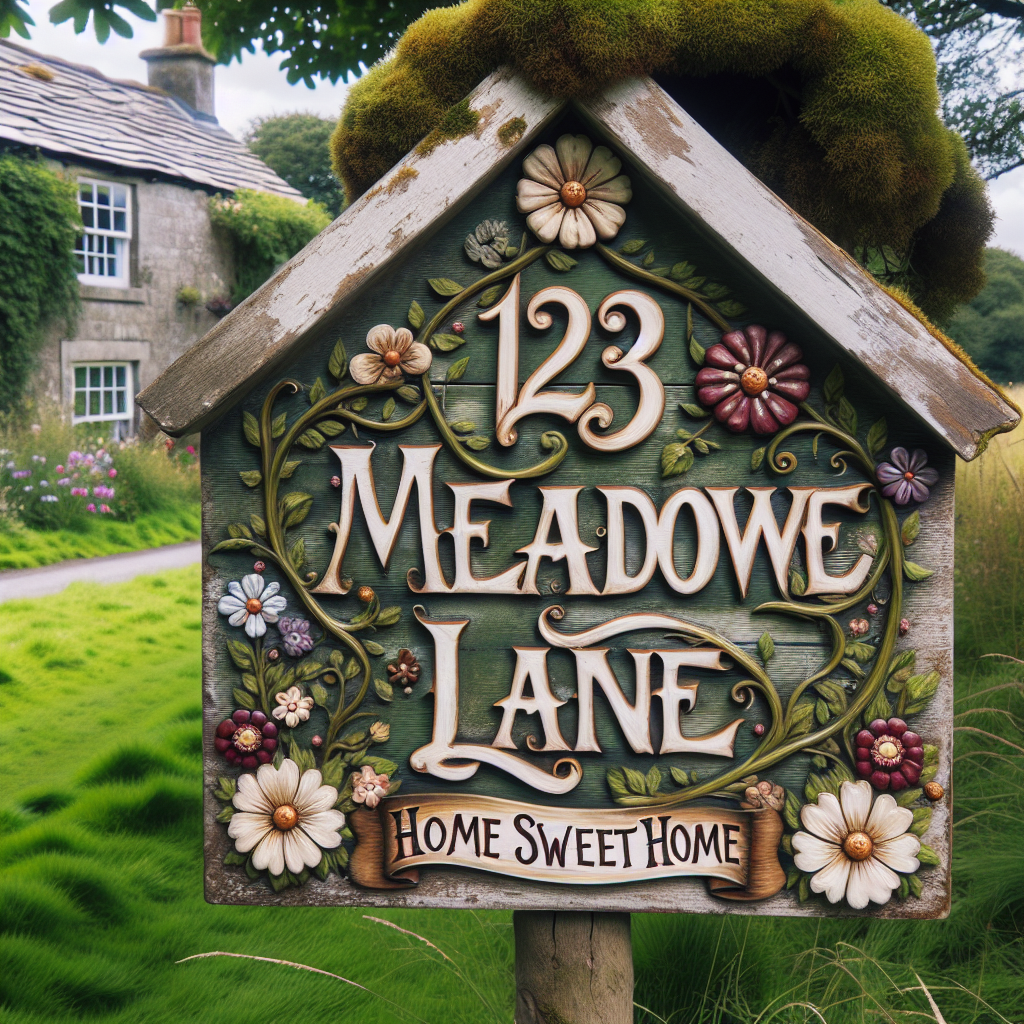} & \includegraphics[width=.5\linewidth,valign=m,height=.2\linewidth]{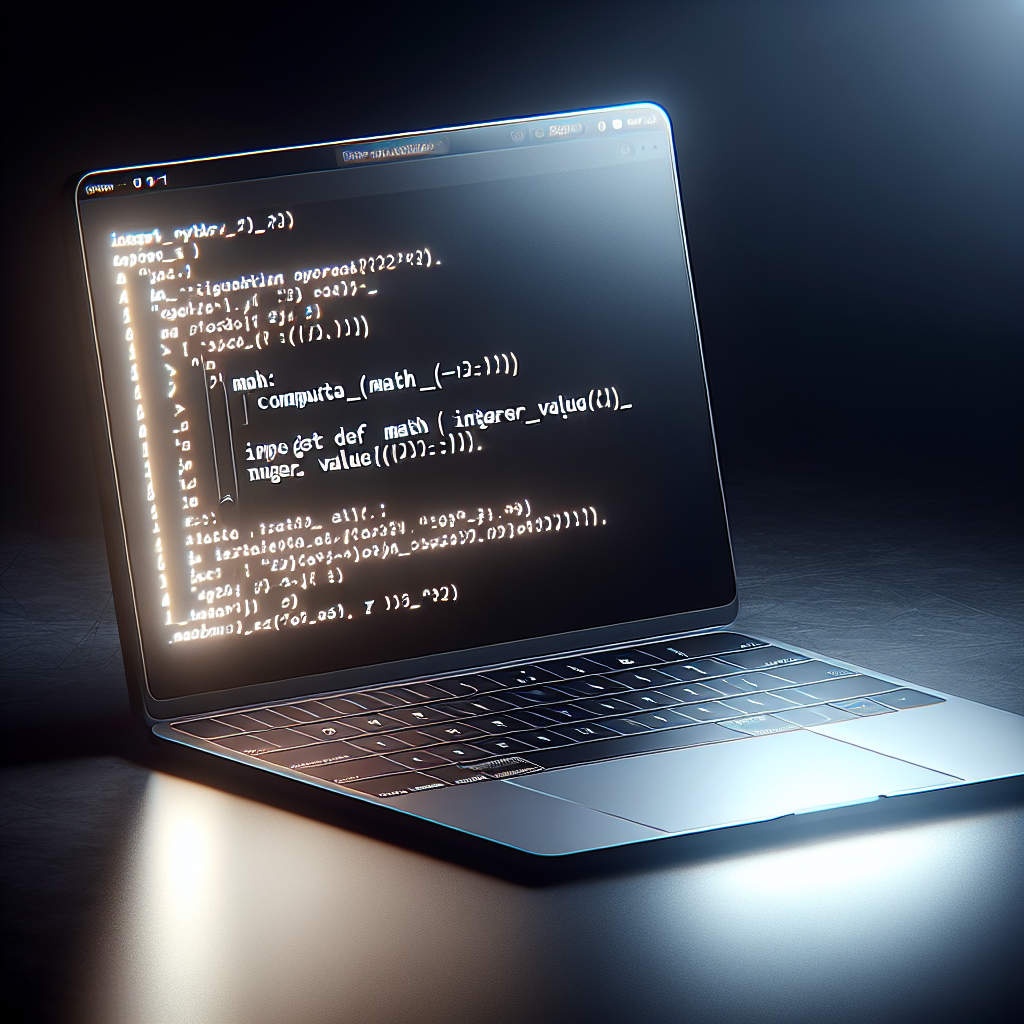} \\ \addlinespace
\end{tabular}
\vspace{-3mm}
\caption{DALLE-3 example generations. On the \textbf{left}, see some generations with higher text fidelity. On the \textbf{right}, see some generations with lower text fidelity.}
\vspace{-20mm}
\end{table*}
\clearpage

\subsection{Stable Diffusion 3}
\begin{table*}[ht!]
\centering
\vspace{-2mm}
\begin{tabular}{ll}
\includegraphics[width=.5\linewidth,valign=m,height=.2\linewidth]{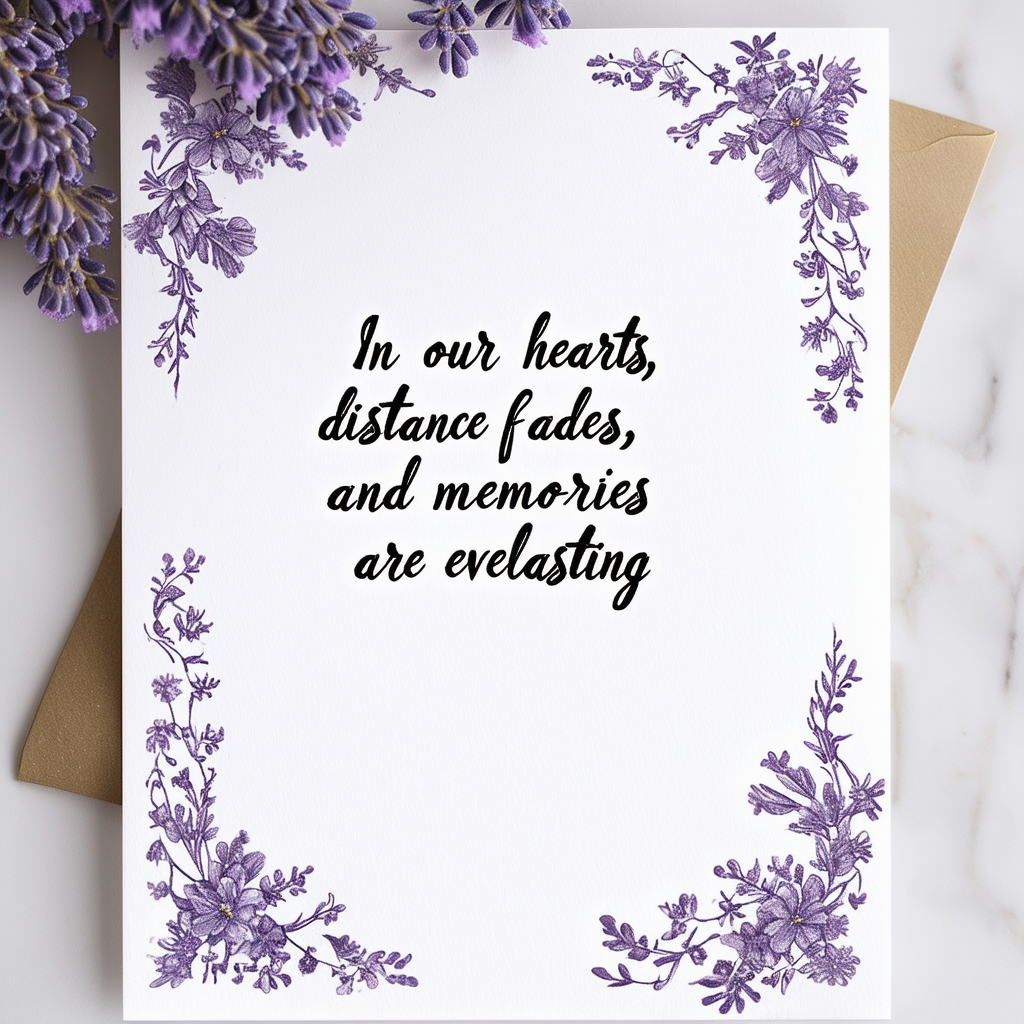} & \includegraphics[width=.5\linewidth,valign=m,height=.2\linewidth]{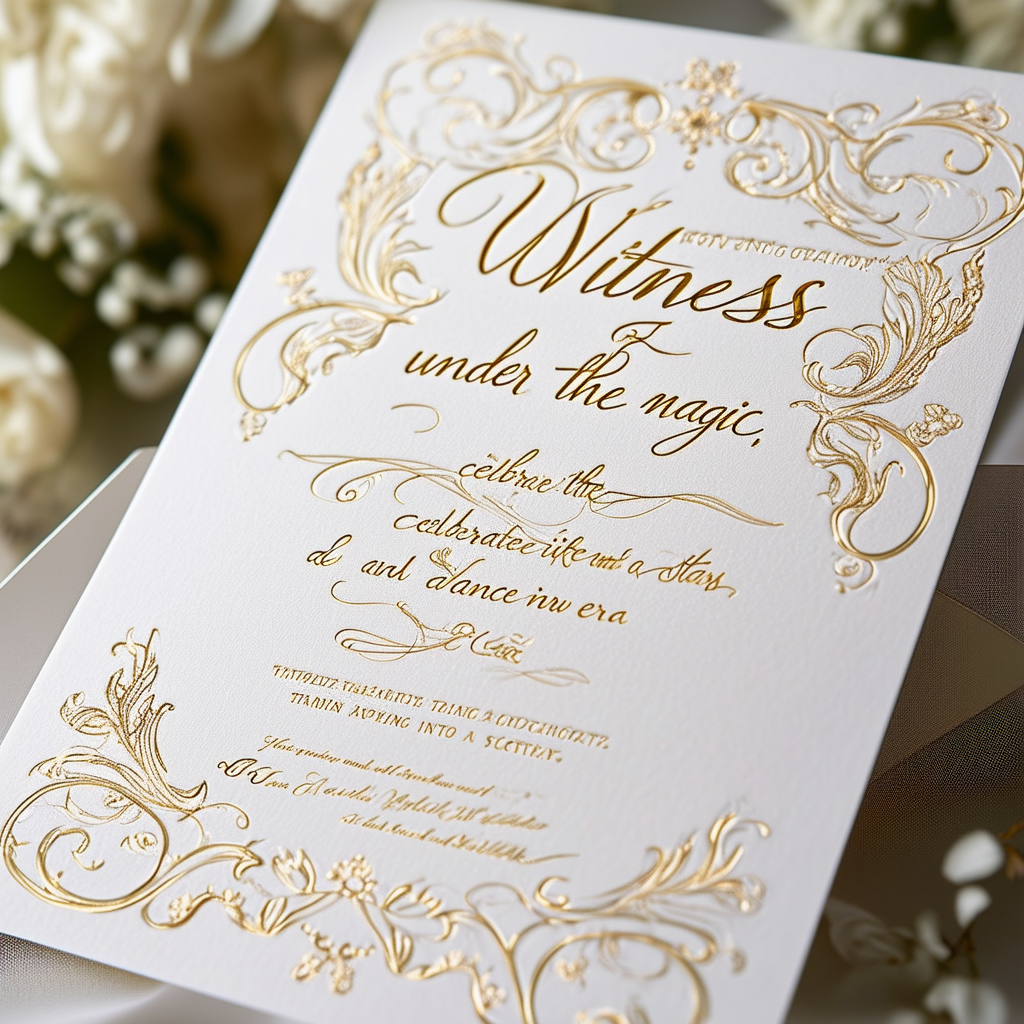} \\ \addlinespace
\includegraphics[width=.5\linewidth,valign=m,height=.2\linewidth]{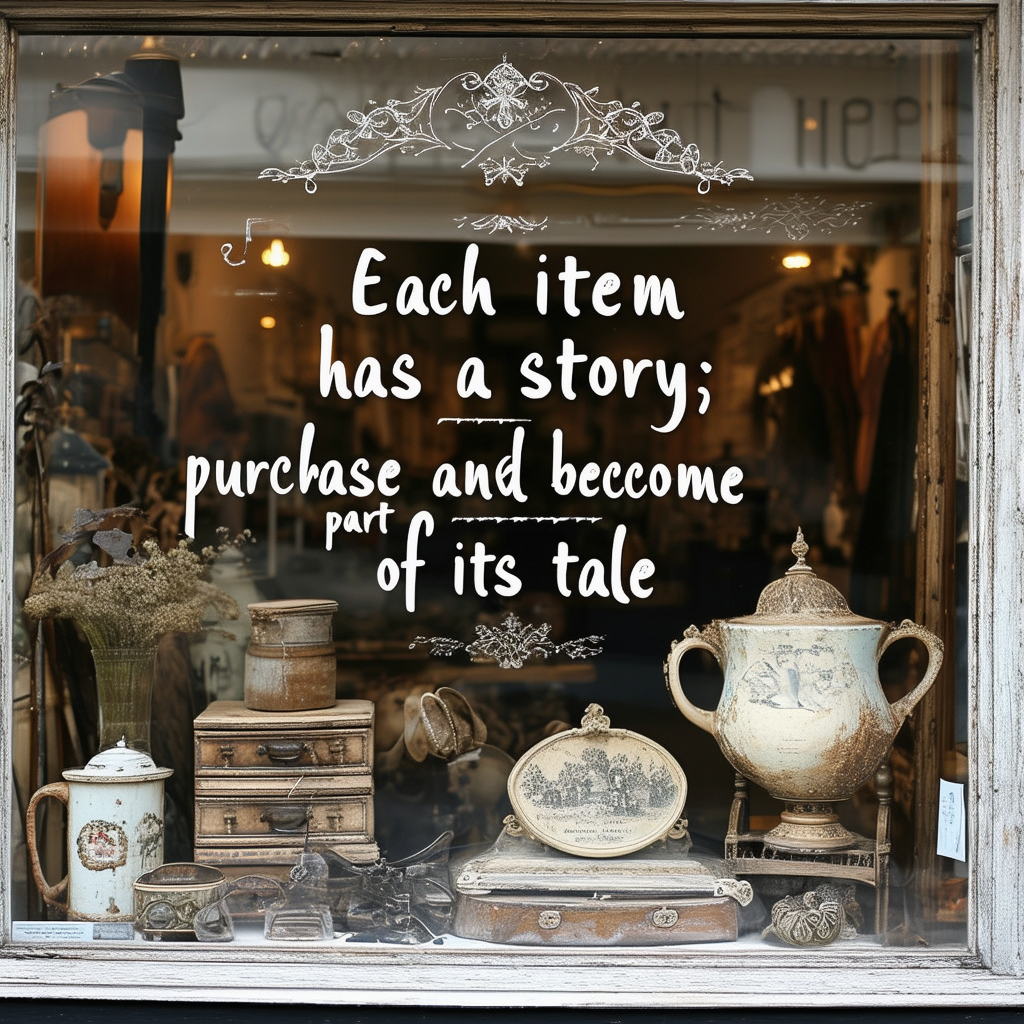} & \includegraphics[width=.5\linewidth,valign=m,height=.2\linewidth]{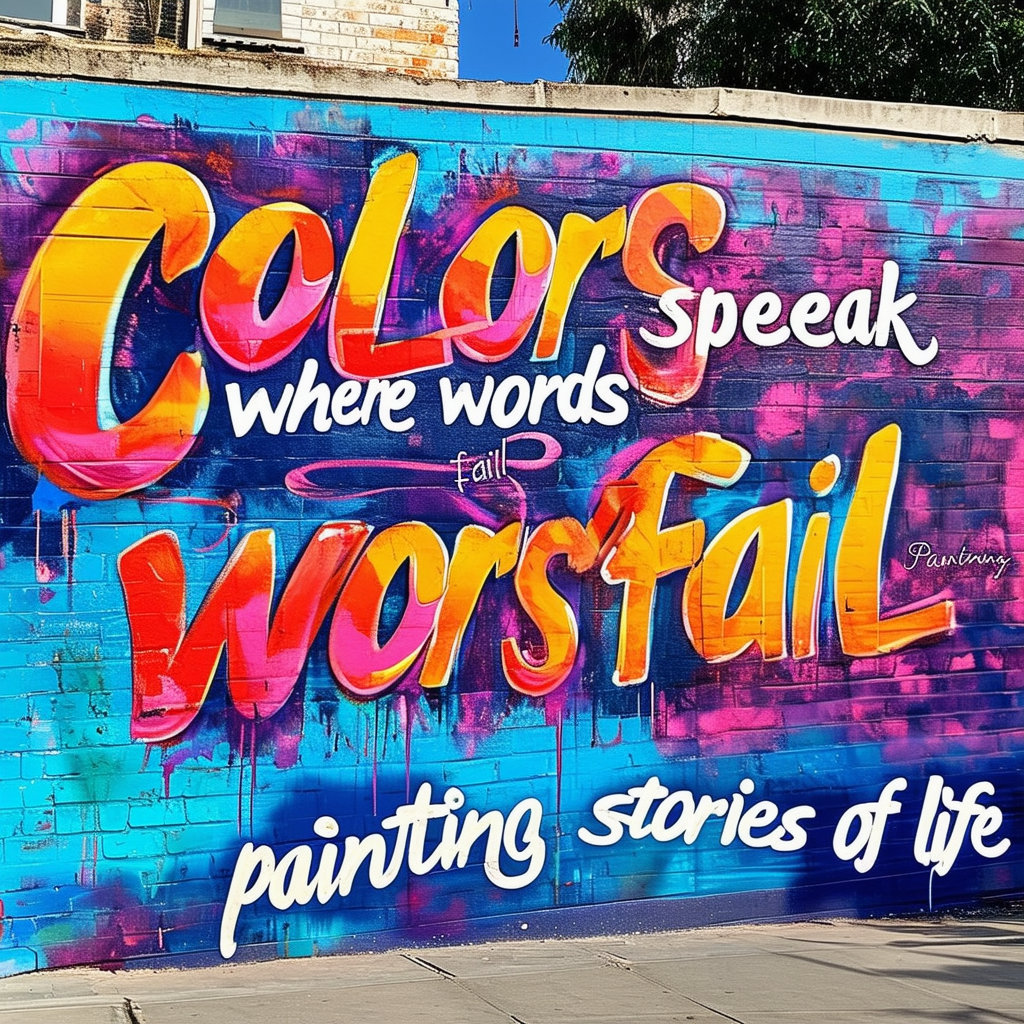} \\ \addlinespace
\includegraphics[width=.5\linewidth,valign=m,height=.2\linewidth]{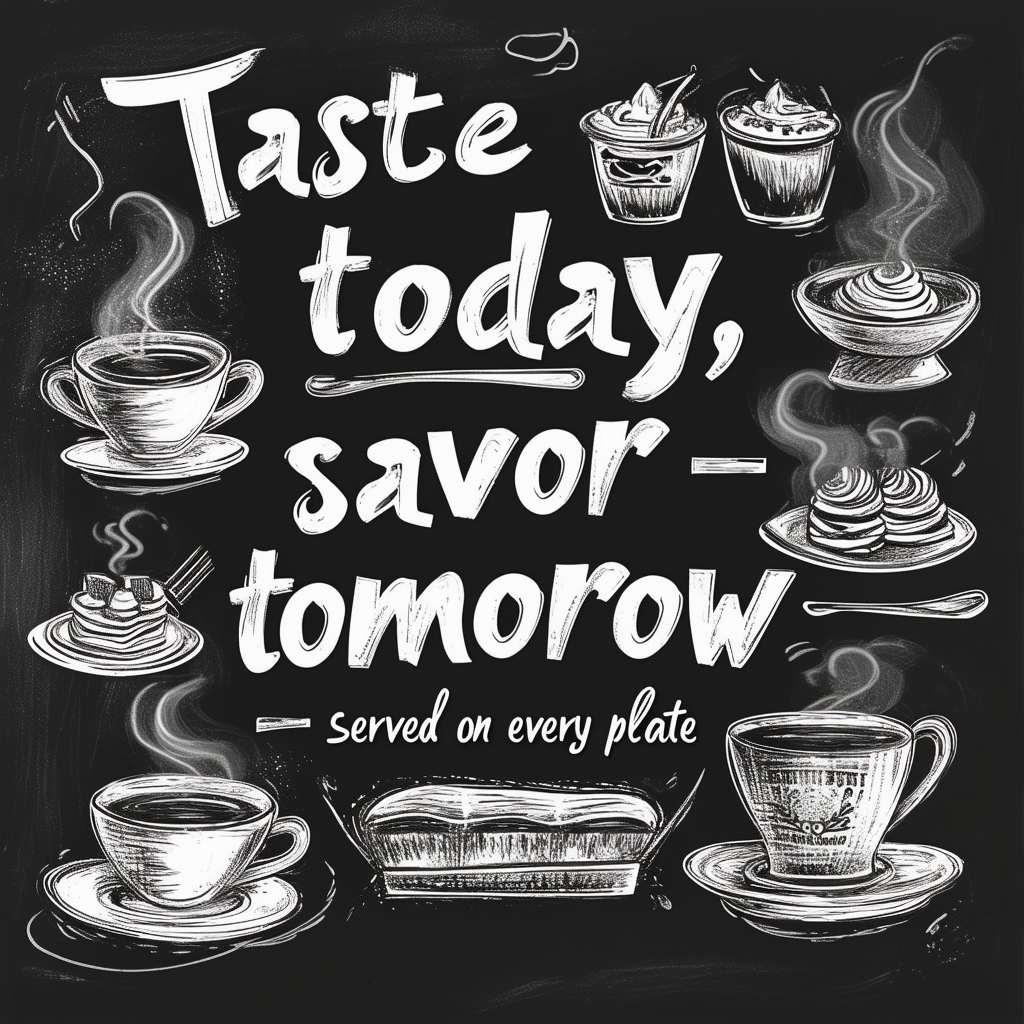} & \includegraphics[width=.5\linewidth,valign=m,height=.2\linewidth]{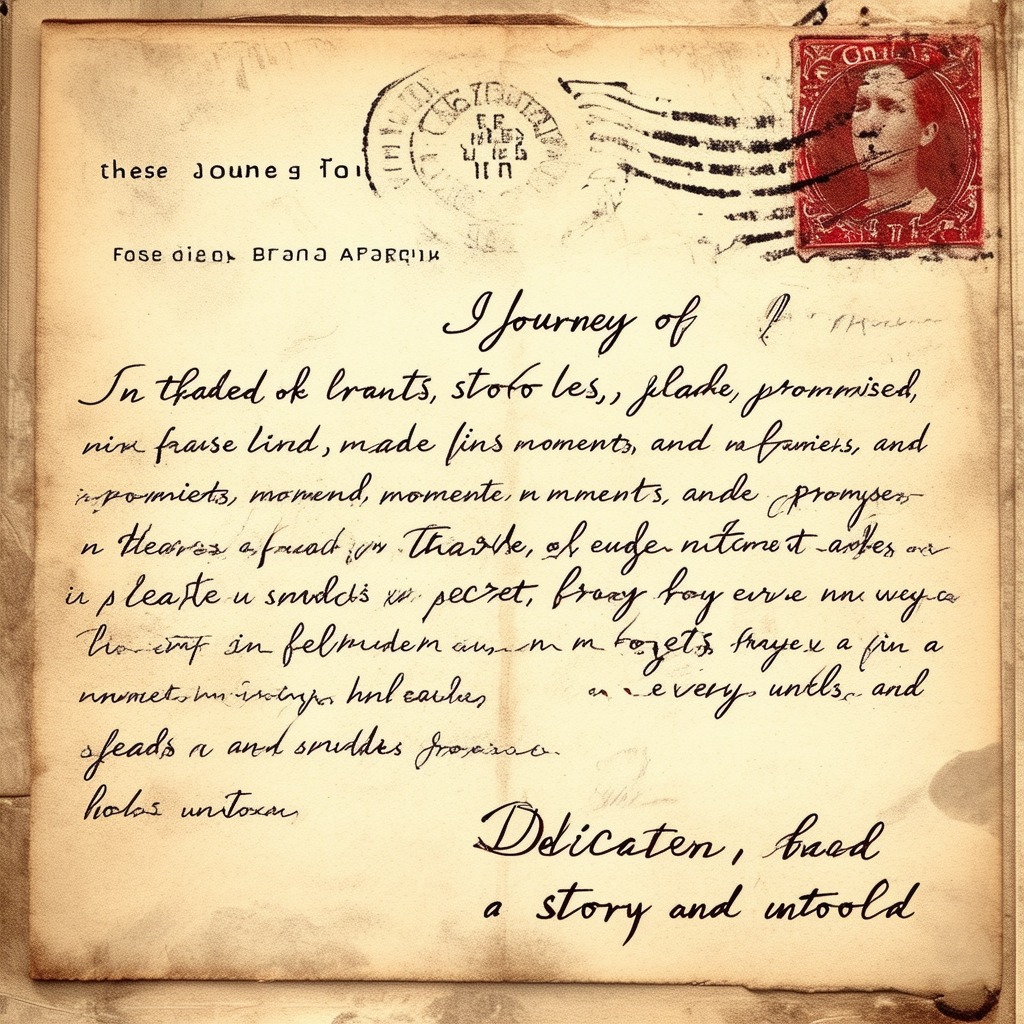} \\ \addlinespace
\includegraphics[width=.5\linewidth,valign=m,height=.2\linewidth]{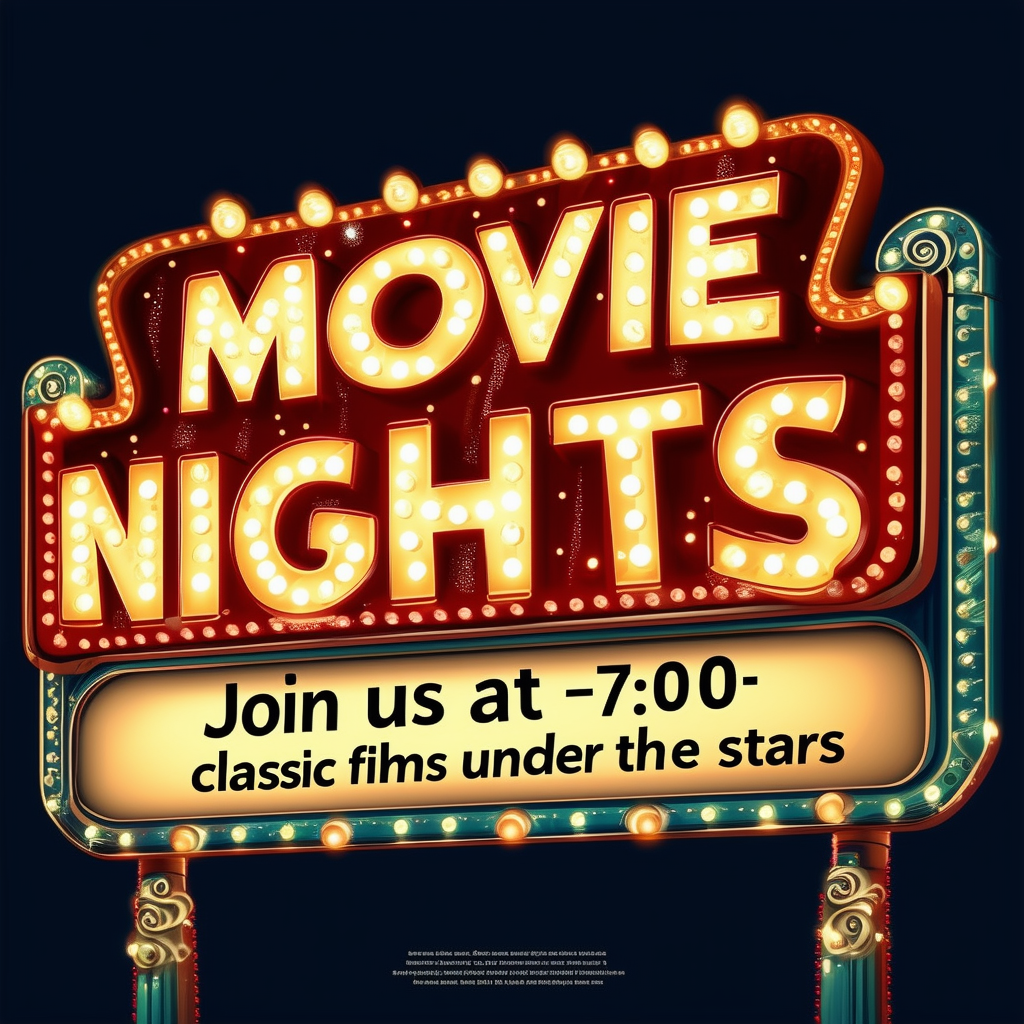} & \includegraphics[width=.5\linewidth,valign=m,height=.2\linewidth]{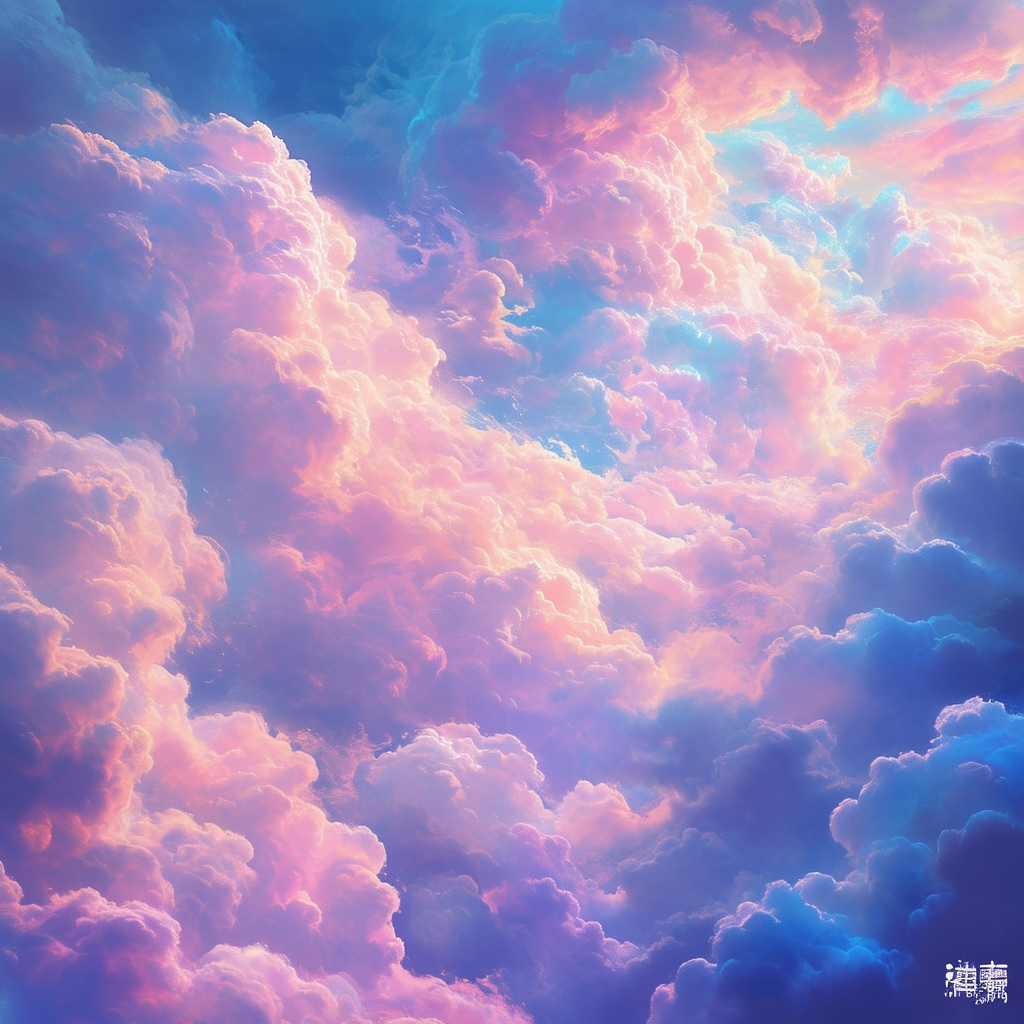} \\ \addlinespace
\end{tabular}
\vspace{-3mm}
\caption{Stable Diffusion 3 example generations. On the \textbf{left}, see some generations with higher text fidelity. On the \textbf{right}, see some generations with lower text fidelity.}
\vspace{-20mm}
\end{table*}
\clearpage

\subsection{ideogram}
\begin{table*}[ht!]
\centering
\vspace{-2mm}
\begin{tabular}{ll}
\includegraphics[width=.5\linewidth,valign=m,height=.2\linewidth]{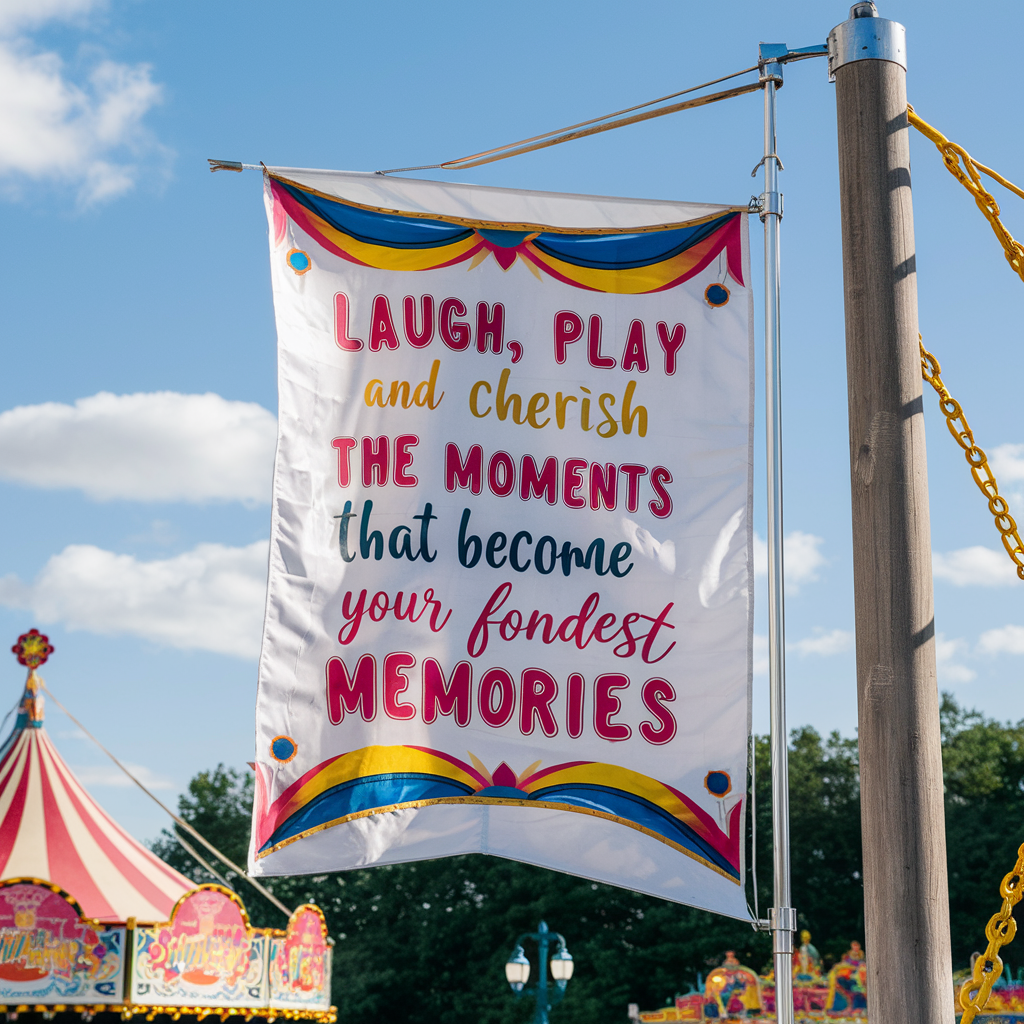} & \includegraphics[width=.5\linewidth,valign=m,height=.2\linewidth]{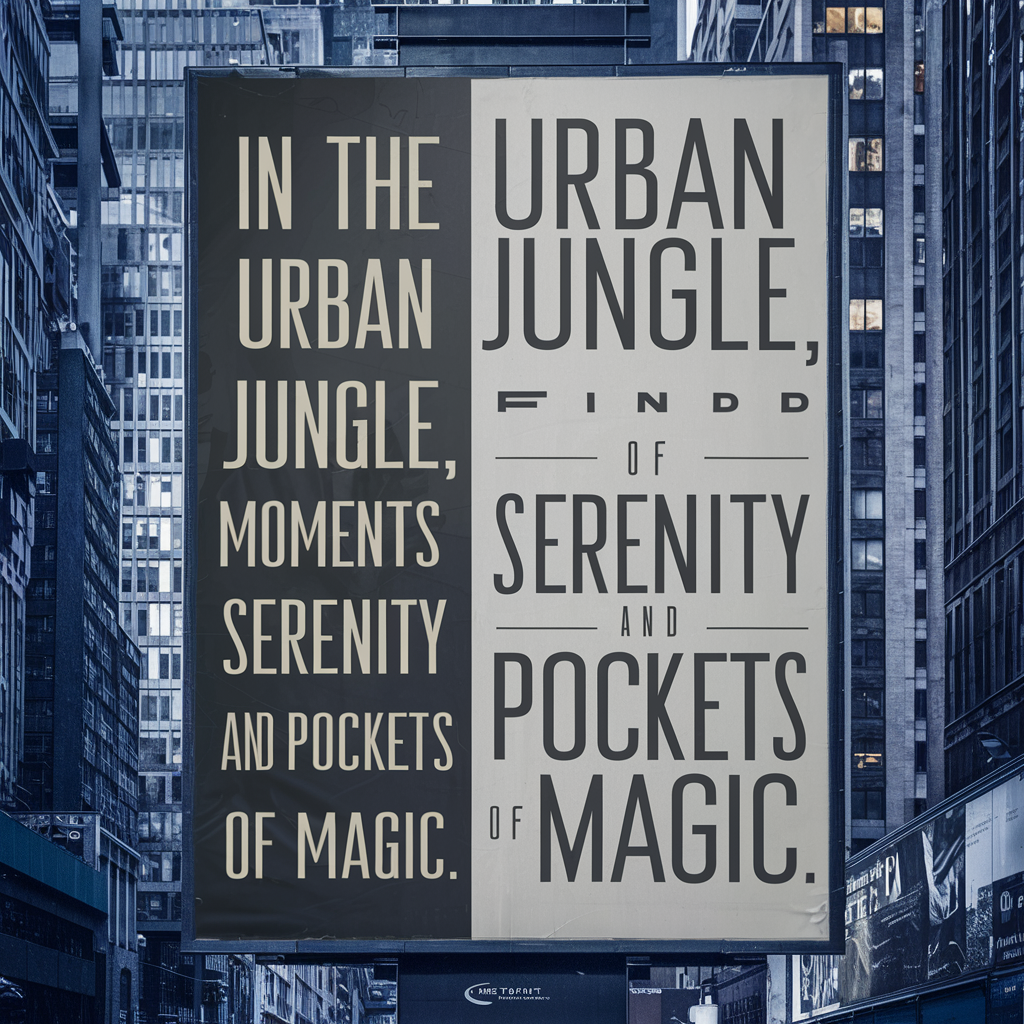} \\ \addlinespace
\includegraphics[width=.5\linewidth,valign=m,height=.2\linewidth]{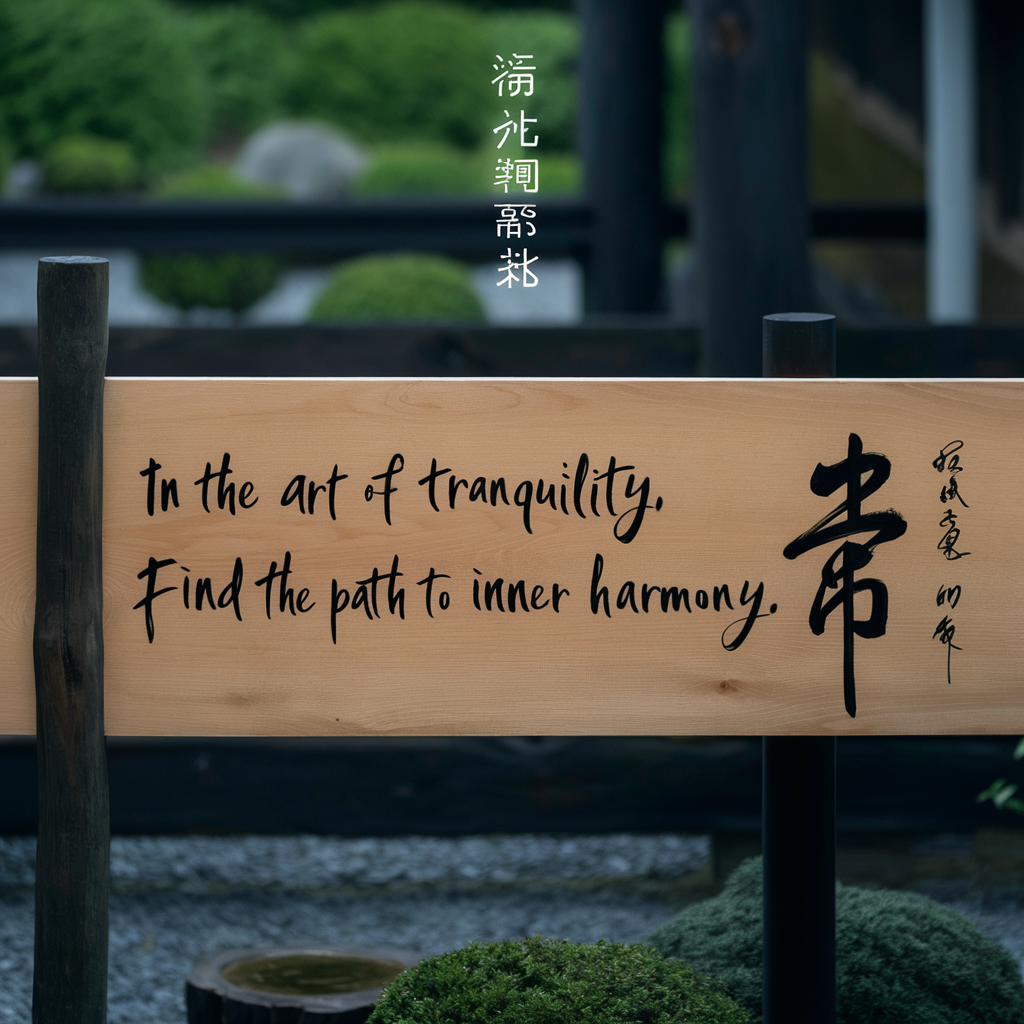} & \includegraphics[width=.5\linewidth,valign=m,height=.2\linewidth]{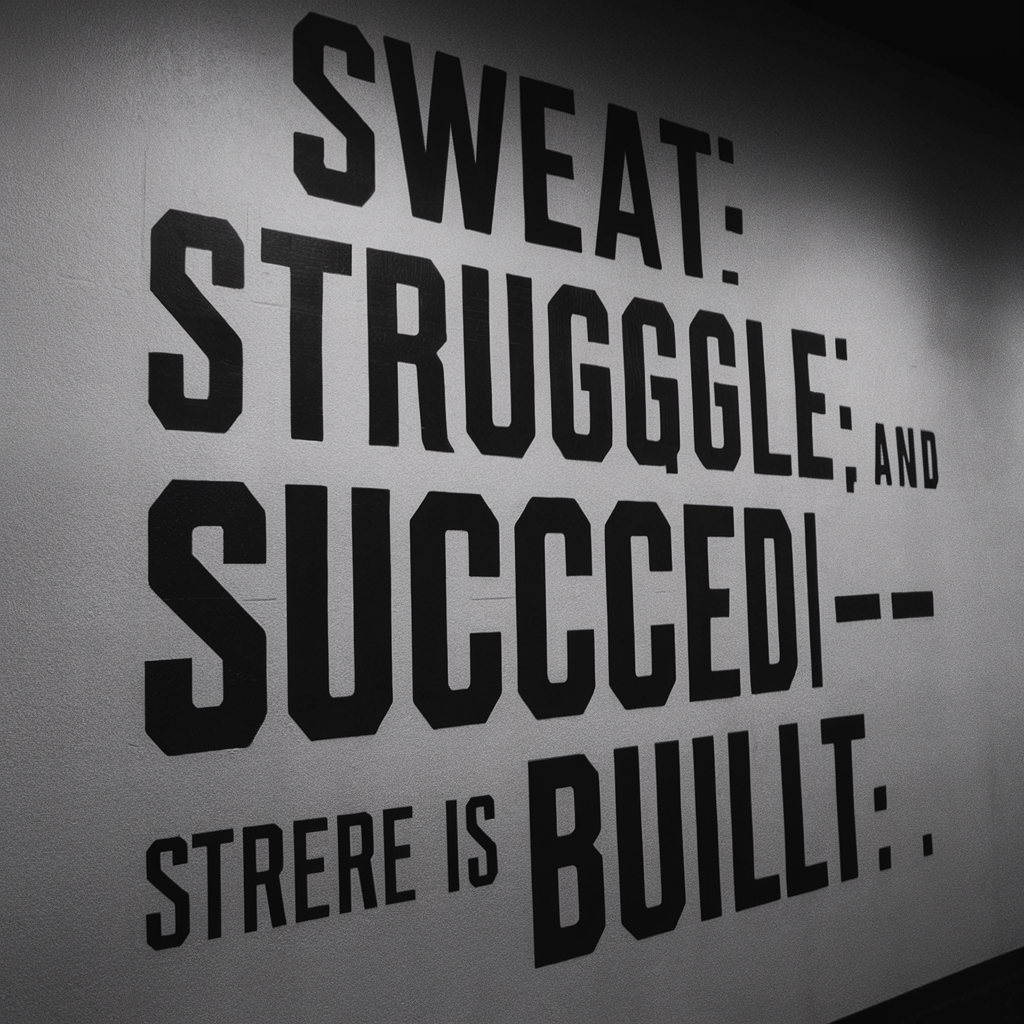} \\ \addlinespace
\includegraphics[width=.5\linewidth,valign=m,height=.2\linewidth]{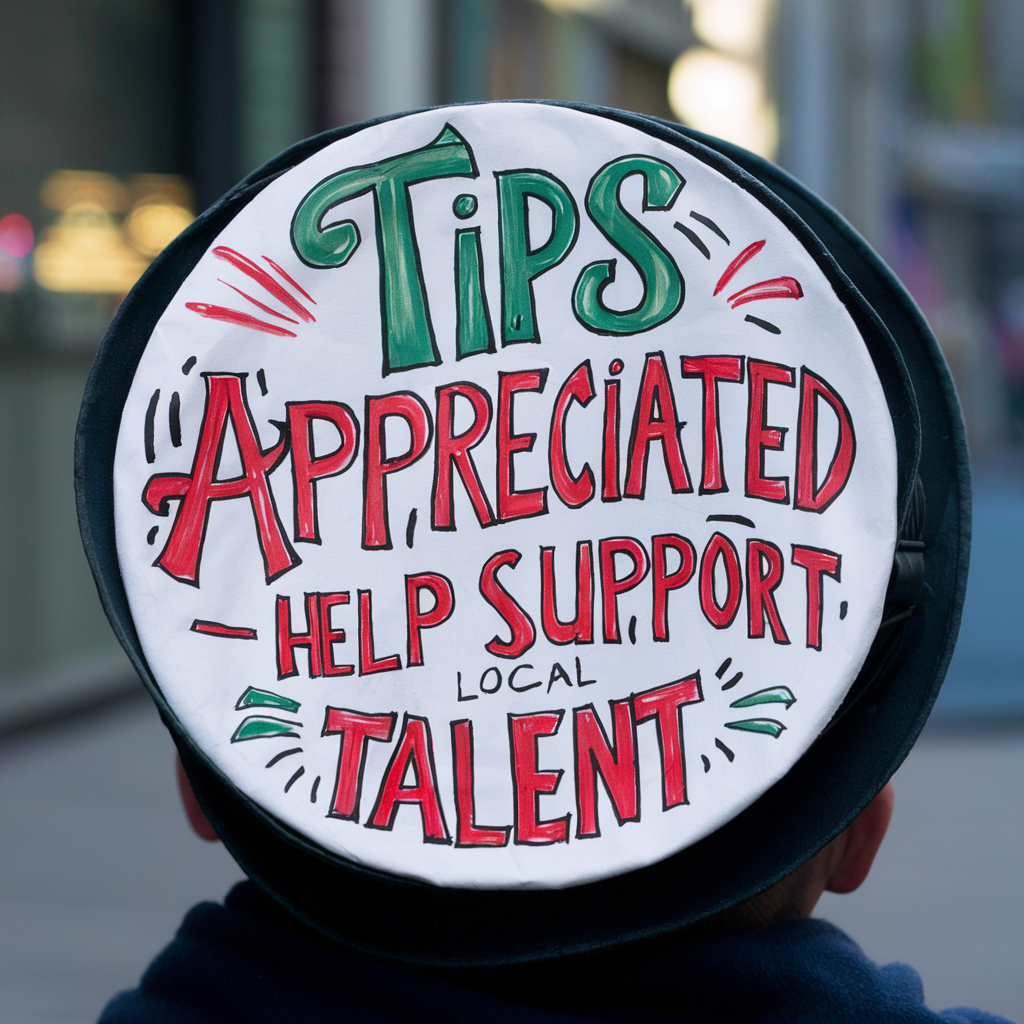} & \includegraphics[width=.5\linewidth,valign=m,height=.2\linewidth]{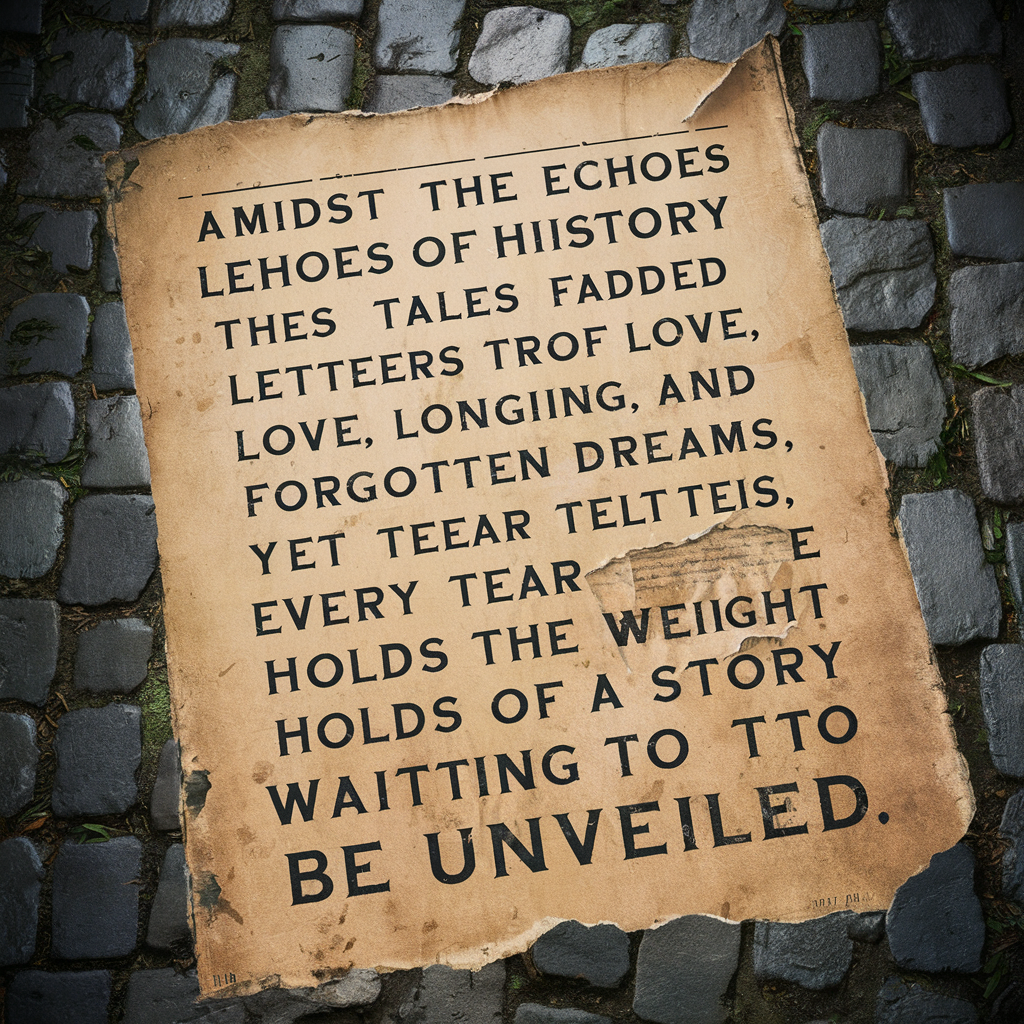} \\ \addlinespace
\includegraphics[width=.5\linewidth,valign=m,height=.2\linewidth]{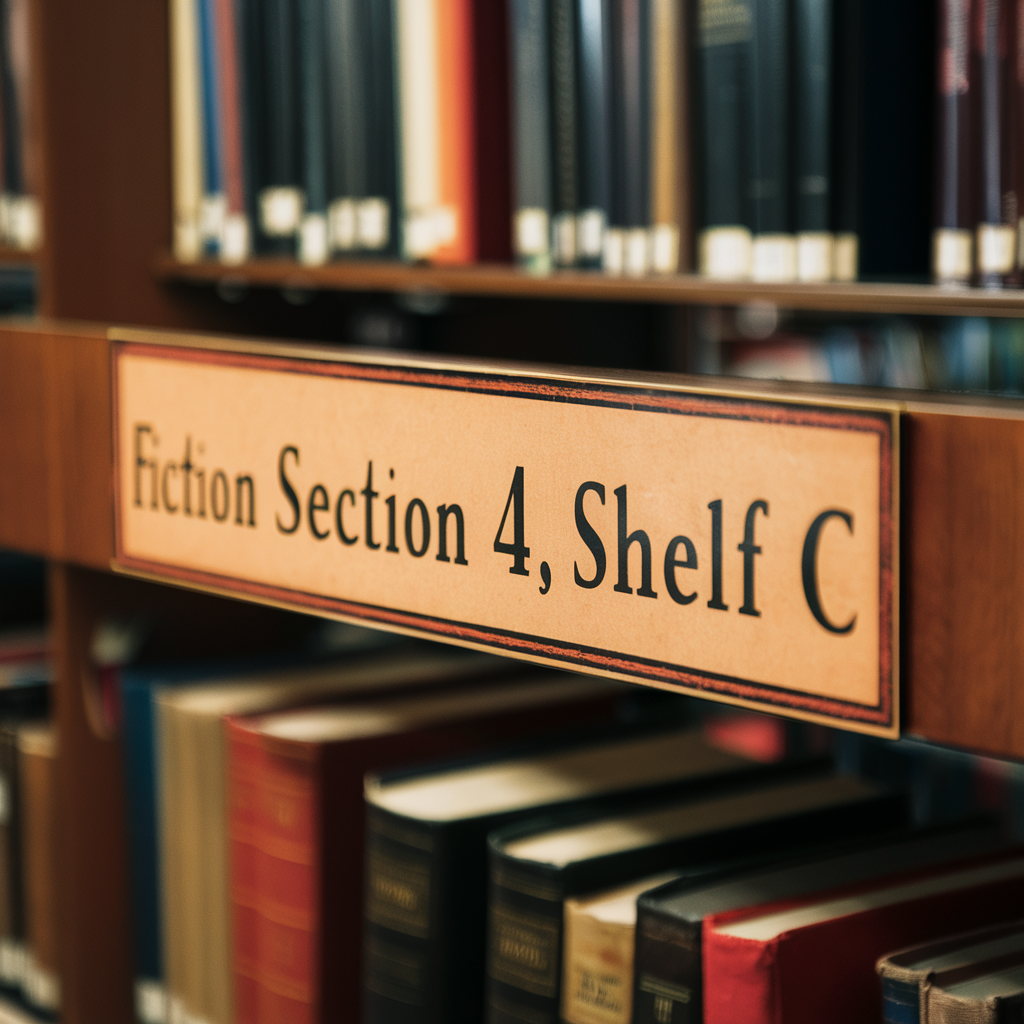} & \includegraphics[width=.5\linewidth,valign=m,height=.2\linewidth]{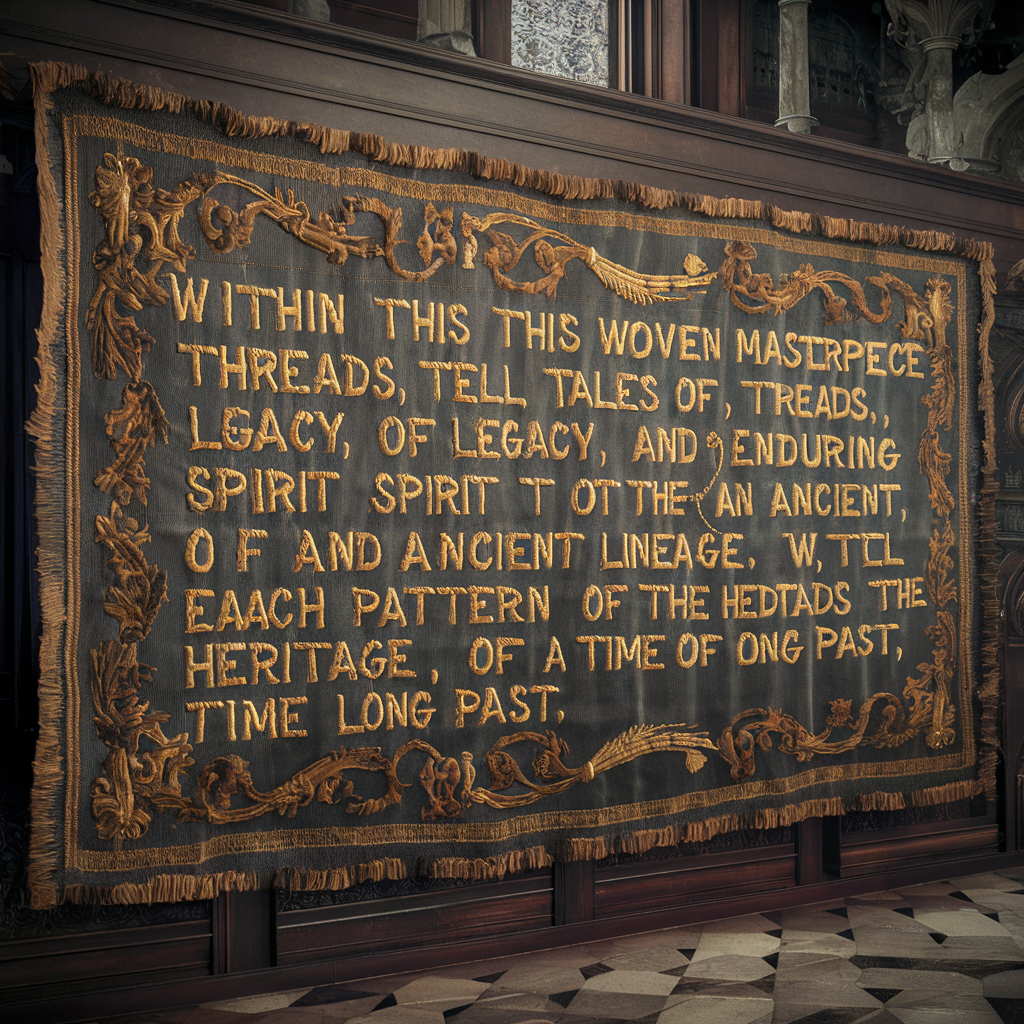} \\ \addlinespace
\end{tabular}
\vspace{-3mm}
\caption{ideogram example generations. On the \textbf{left}, see some generations with higher text fidelity. On the \textbf{right}, see some generations with lower text fidelity.}
\vspace{-20mm}
\end{table*}

%
%
%
%